\documentclass[runningheads]{llncs}
\usepackage{graphicx}

\usepackage{tikz}
\usepackage{comment}
\usepackage{amsmath,amssymb} 
\usepackage{color}
\usepackage{booktabs}
\usepackage{caption}
\usepackage{subcaption}
\usepackage{siunitx}
\usepackage{hyperref}
\usepackage{booktabs}
\usepackage{wrapfig,lipsum}

\DeclareMathOperator*{\argmin}{arg\,min}
\newcommand{\el}[1]{#1~\textit{et al.}}
\newcommand\figcaption{\def\@captype{figure}\caption}
\newcommand\tabcaption{\def\@captype{table}\caption}
\newlength{\oldintextsep}
\usepackage[accsupp]{axessibility}  

\begin{document}
\pagestyle{headings}
\mainmatter
\def\ECCVSubNumber{4547}  

\title{Bringing Rolling Shutter Images Alive with Dual Reversed Distortion} 

\titlerunning{Dual Reversed RS}
%
\author{Zhihang Zhong\inst{1,4} \and
Mingdeng Cao\inst{2} \and
Xiao Sun\inst{3} \and
Zhirong Wu\inst{3} \and
Zhongyi Zhou\inst{1}\and
Yinqiang Zheng\inst{1} \and
Stephen Lin\inst{3} \and
Imari Sato\inst{1,4}}
\authorrunning{Z. Zhong et al.}
%
\institute{The University of Tokyo, \email{zhong@is.s.u-tokyo.ac.jp} \and
Tsinghua University \and
Microsoft Research Asia \and
National Institute of Informatics}
\maketitle

\begin{abstract}
Rolling shutter (RS) distortion can be interpreted as the result of picking a row of pixels from instant global shutter (GS) frames over time during the exposure of the RS camera. This means that the information of each instant GS frame is partially, yet sequentially, embedded into the row-dependent distortion. Inspired by this fact, we address the challenging task of reversing this process, \textit{i.e.}, extracting undistorted GS frames from images suffering from RS distortion. However, since RS distortion is coupled with other factors such as readout settings and the relative velocity of scene elements to the camera, models that only exploit the geometric correlation between temporally adjacent images suffer from poor generality in processing data with different readout settings and dynamic scenes with both camera motion and object motion. In this paper, instead of two consecutive frames, we propose to exploit a pair of images captured by dual RS cameras with reversed RS directions for this highly challenging task. Grounded on the symmetric and complementary nature of dual reversed distortion, we develop a novel end-to-end model, IFED, to generate dual optical flow sequence through iterative learning of the velocity field during the RS time. Extensive experimental results demonstrate that IFED is superior to naive cascade schemes, as well as the state-of-the-art which utilizes adjacent RS images. Most importantly, although it is trained on a synthetic dataset, IFED is shown to be effective at retrieving GS frame sequences from real-world RS distorted images of dynamic scenes. Code is available at \href{https://github.com/zzh-tech/Dual-Reversed-RS}{https://github.com/zzh-tech/Dual-Reversed-RS}.
\end{abstract}

\section{Introduction}

\begin{figure*}[!t]
  \centering
  \includegraphics[width=\linewidth]{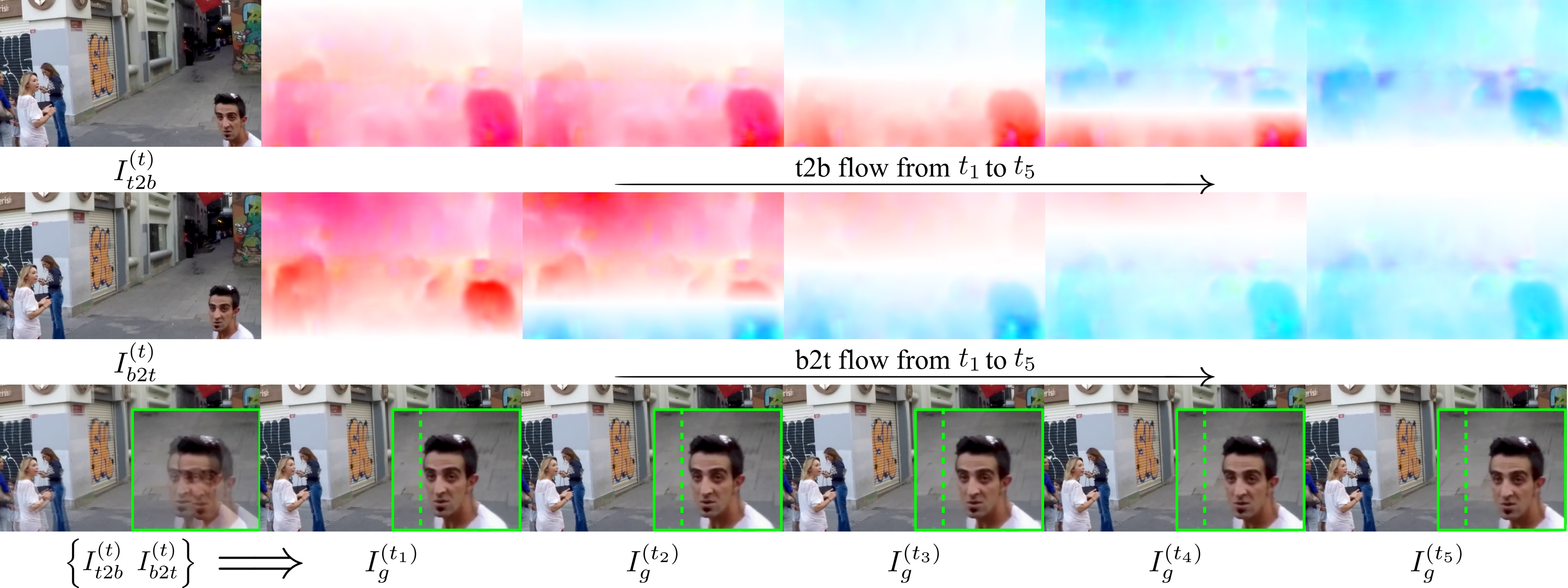}
  \caption{\textbf{Consecutive distortion-free frames extracted from a pair of images with reversed rolling shutter distortion.} The $1^{st}$ row presents the distorted image $I_{t2b}^{(t)}$ from top-to-bottom scanning at time $t$ and the generated optical flows to the extracted frames. The $2^{nd}$ row presents the distorted image $I_{b2t}^{(t)}$ from bottom-to-top scanning at the same time and its corresponding optical flows. The $3^{rd}$ row presents the mixed input  $\{I_{t2b}^{(t)}~ I_{b2t}^{(t)}\}$ and the extracted global shutter frames $\{I_{g}^{(t_i)}\}$ in chronological order.}
  \label{fig:teaser}
\end{figure*}

Rolling shutter (RS) cameras are used in many devices such as smartphones and self-driving vision systems due to their low cost and high data transfer rate~\cite{litwiller2001ccd}. Compared to global shutter (GS) cameras, which capture the whole scene at a single instant, RS cameras scan the scene row-by-row to produce an image. This scanning mechanism may be viewed as sub-optimal because it leads to RS distortion, also known as the jello effect, in the presence of camera and/or object motion. However, we argue that RS photography encodes rich temporal information through its push-broom scanning process. This property provides a critical cue for predicting a sequence of GS images, where distorted images are brought alive at a higher frame rate, which goes beyond the task of recovering a single snapshot as in the RS correction task~\cite{baker2010removing,zhuang2019learning,liu2020deep}, as shown in Fig.~\ref{fig:teaser}.

\begin{figure*}[!t]
	\centering
	\begin{subfigure}[b]{0.3\textwidth}
		\centering
		\includegraphics[width=\textwidth]{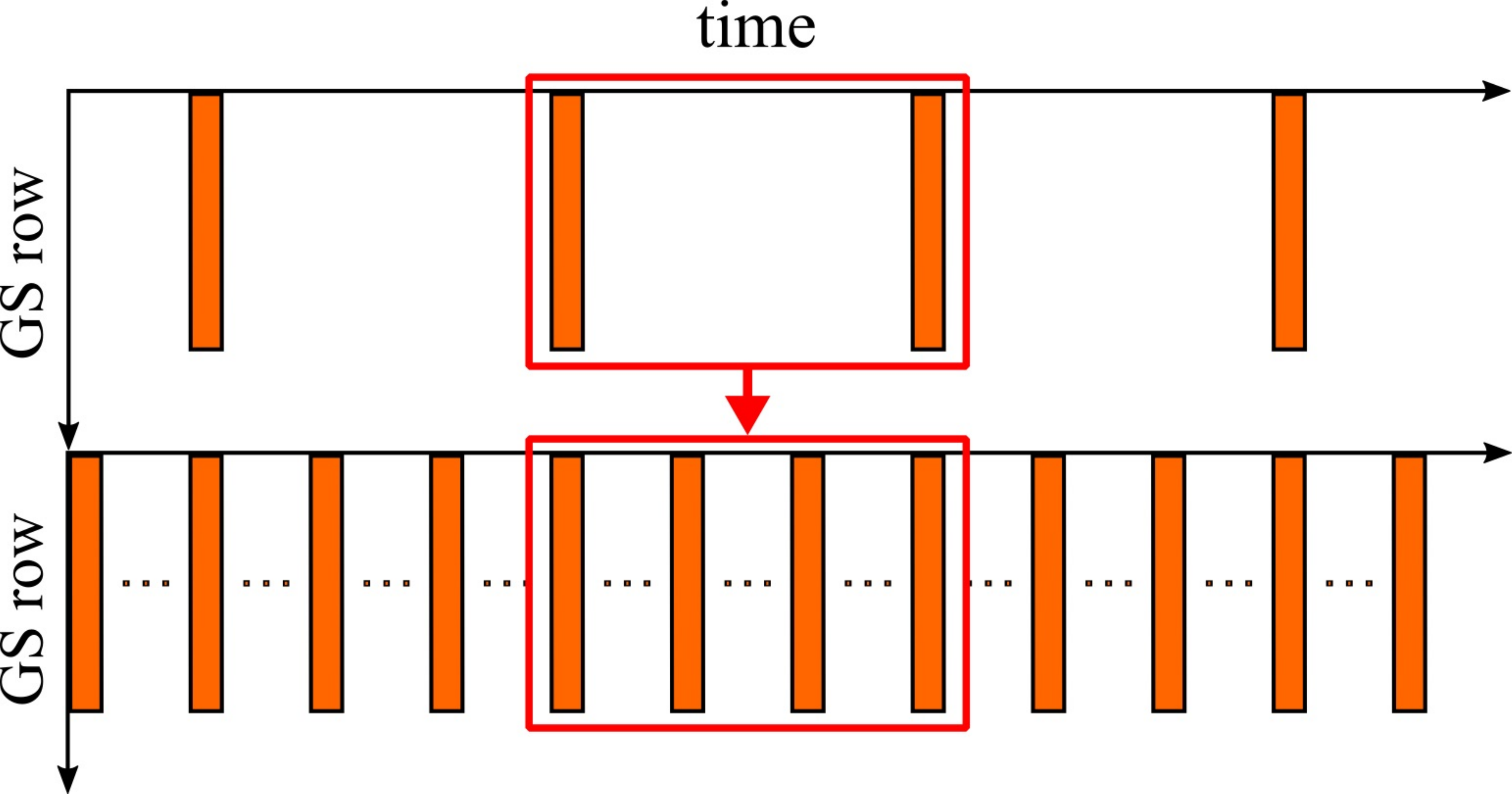}
		\caption{VFI}
		\label{fig:gs2mgs}
	\end{subfigure}
	\hfill
	\begin{subfigure}[b]{0.3\textwidth}
		\centering
		\includegraphics[width=\textwidth]{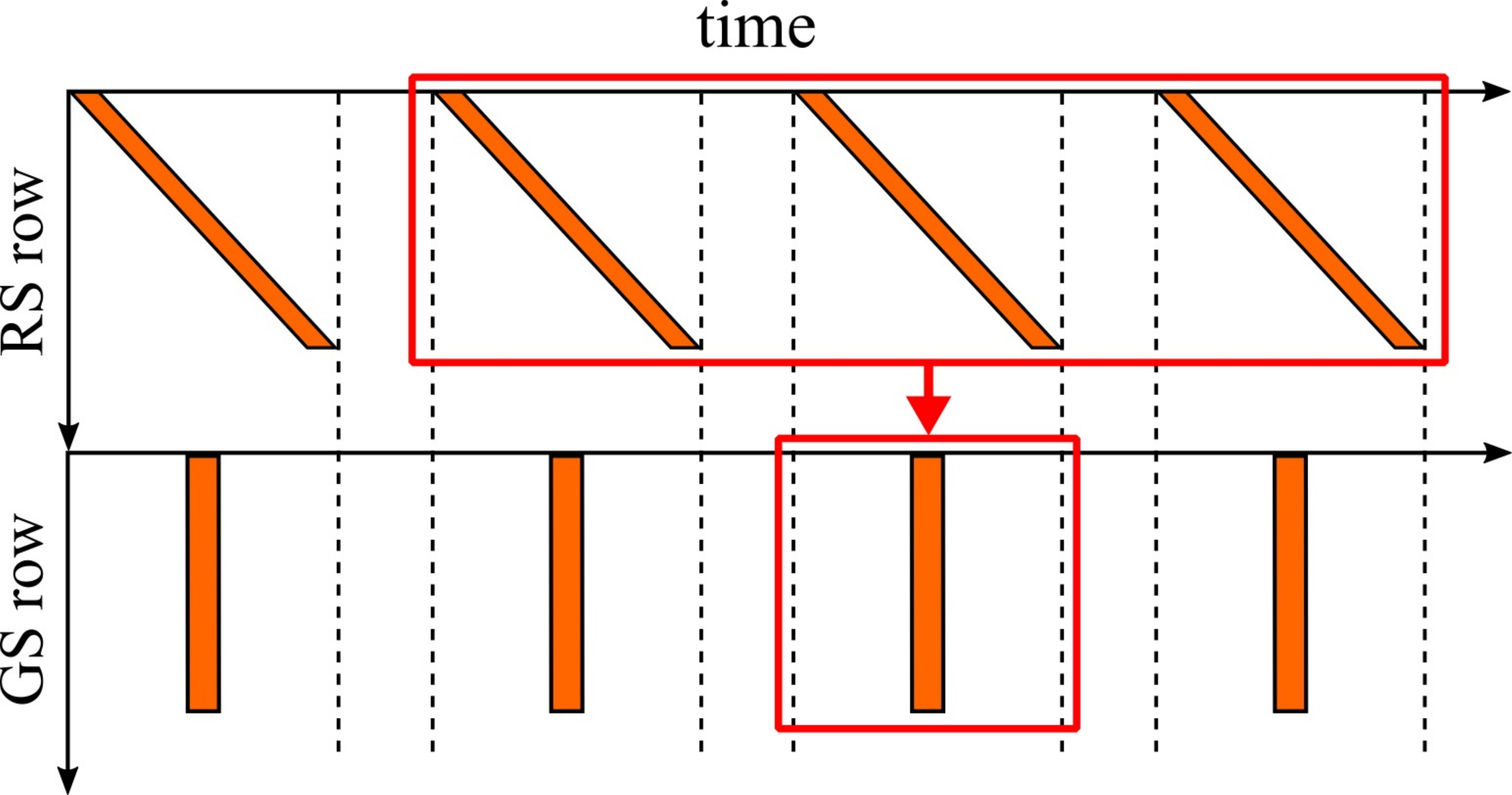}
		\caption{RSC}
		\label{fig:rs2gs}
	\end{subfigure}
	\hfill
	\begin{subfigure}[b]{0.3\textwidth}
		\centering
		\includegraphics[width=\textwidth]{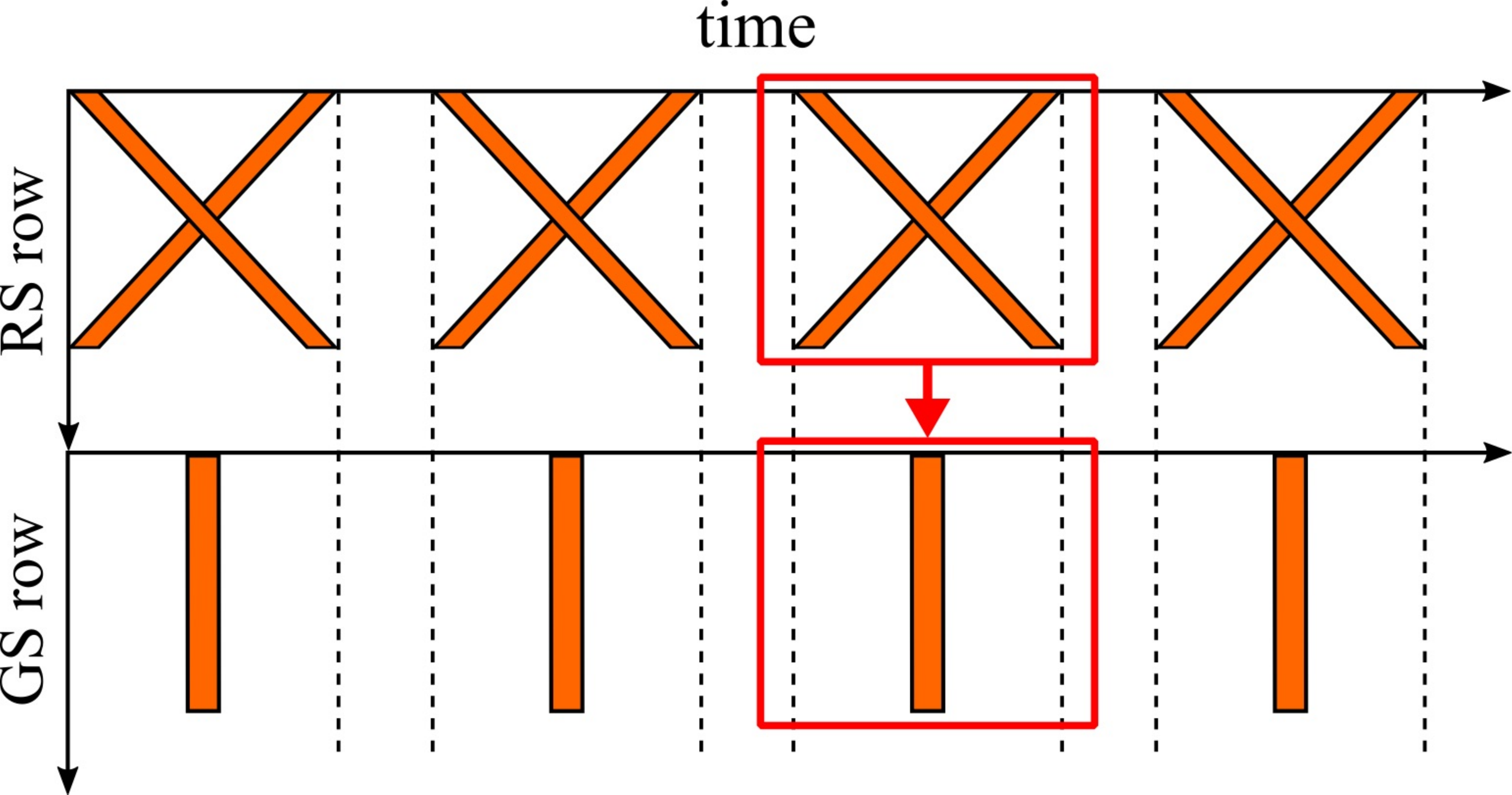}
		\caption{Dual-RSC}
		\label{fig:2rs2gs}
	\end{subfigure}
	\hfill
	\begin{subfigure}[b]{0.3\textwidth}
		\centering
		\includegraphics[width=\textwidth]{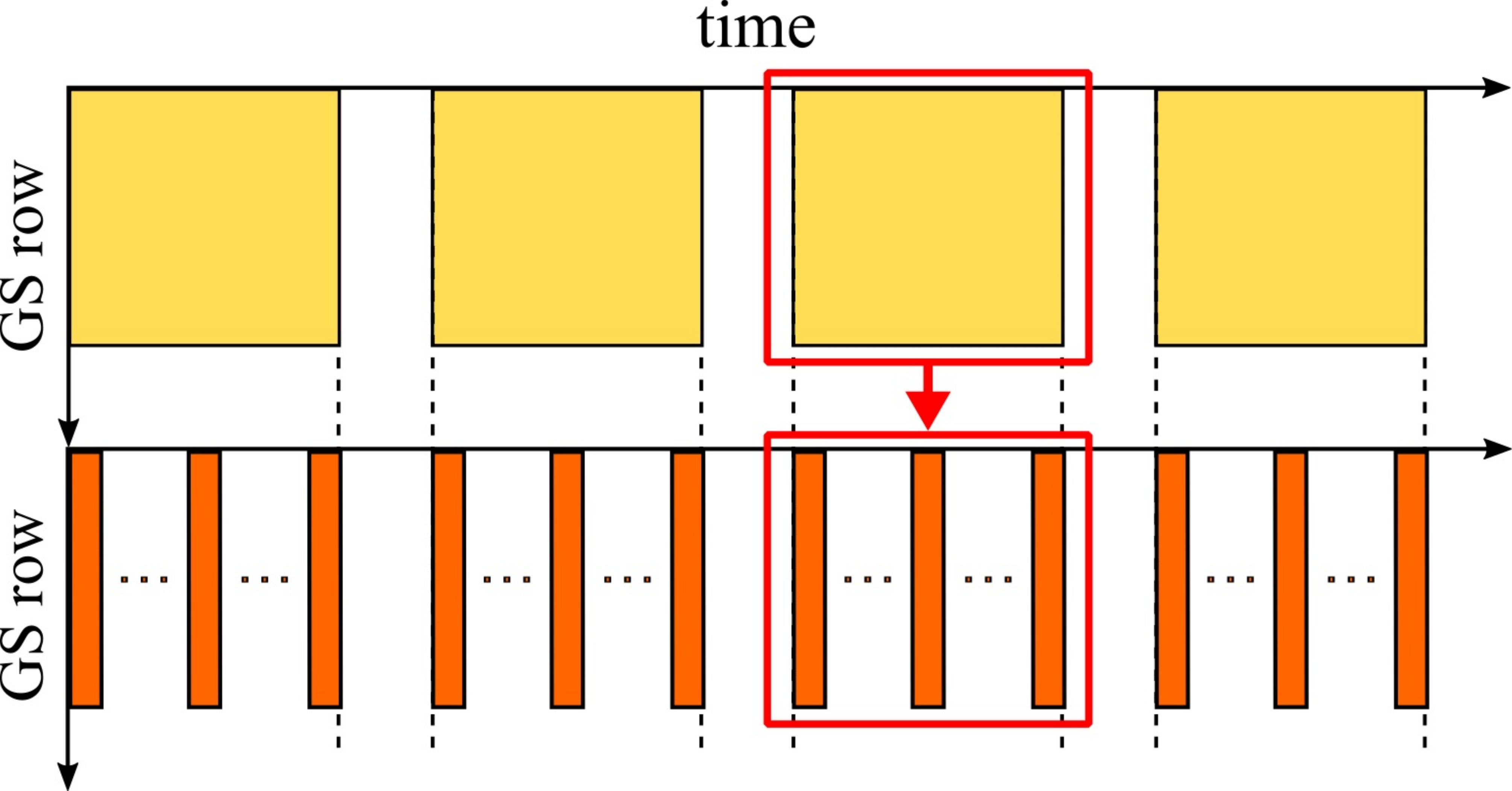}
		\caption{BFI}
		\label{fig:bgs2mgs}
	\end{subfigure}
	\hfill
	\begin{subfigure}[b]{0.3\textwidth}
		\centering
		\includegraphics[width=\textwidth]{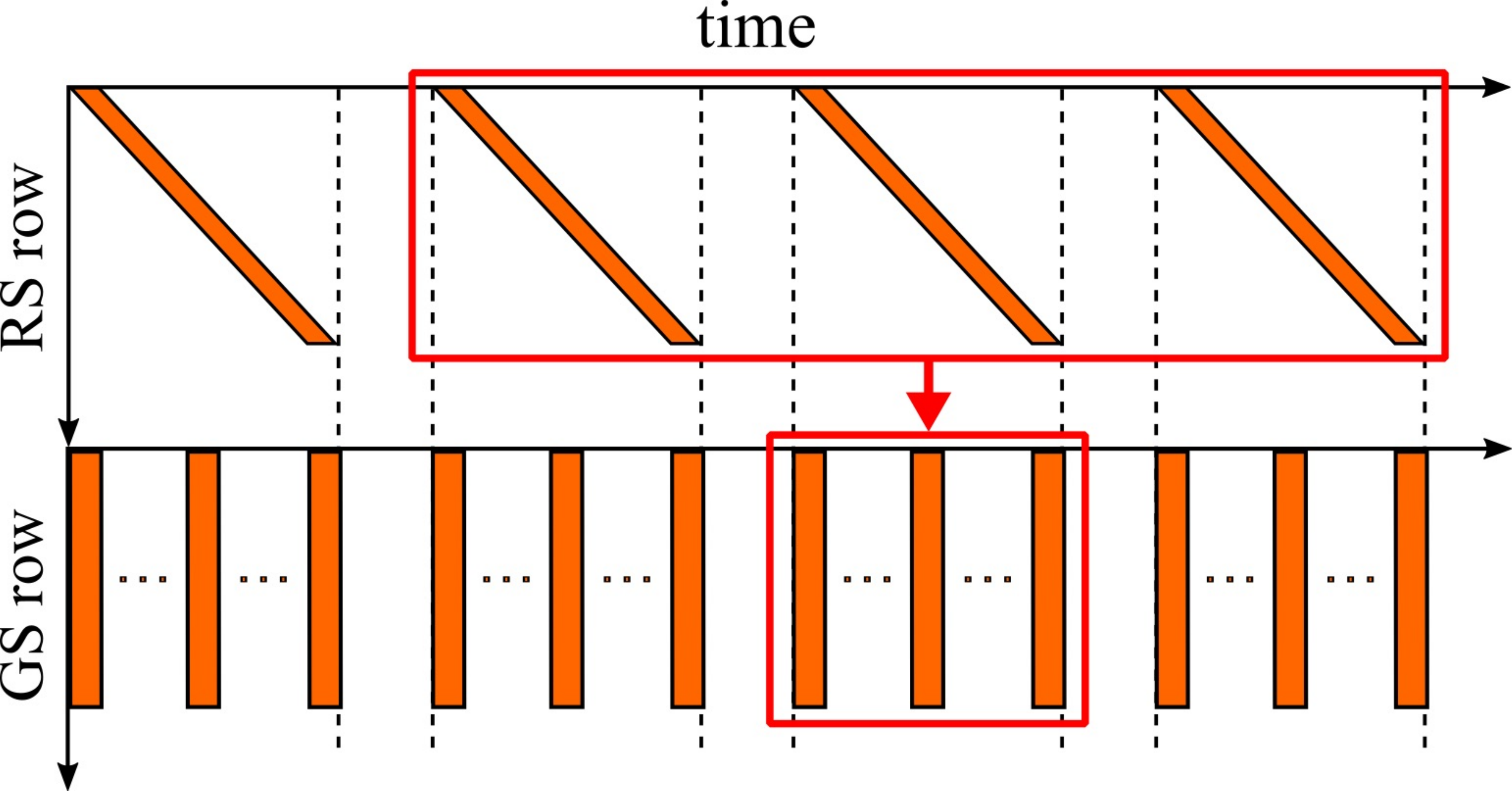}
		\caption{RSCI}
		\label{fig:rs2mgs}
	\end{subfigure}
	\hfill
	\begin{subfigure}[b]{0.3\textwidth}
		\centering
		\includegraphics[width=\textwidth]{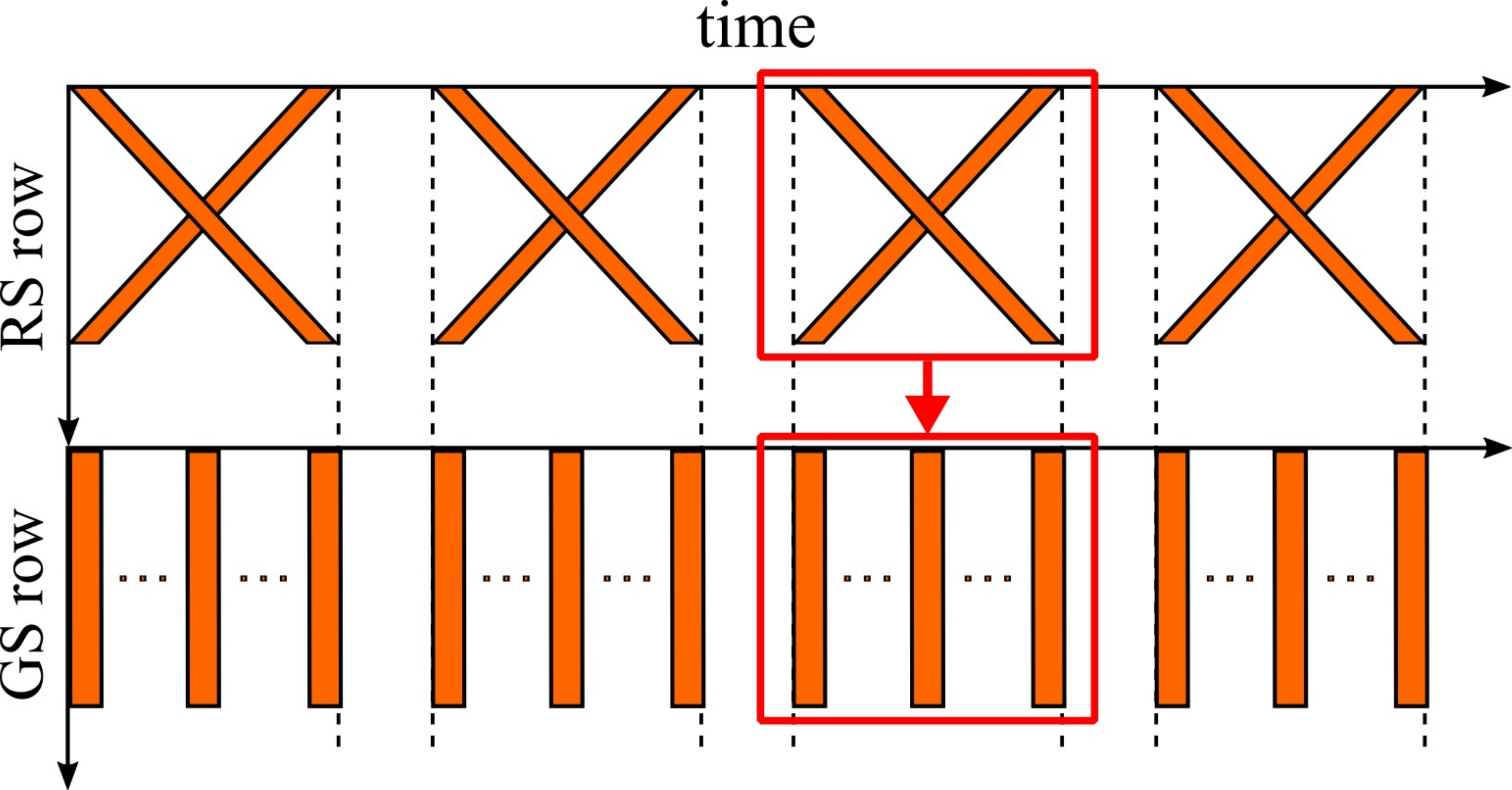}
		\caption{Dual-RSCI}
		\label{fig:2rs2mgs}
	\end{subfigure}
	\caption{\textbf{Comparison of different tasks.}
		The first row represents the input and the second row represents the output of each task. The x-axis and y-axis represent the time and the row location of the captured or generated image, respectively. (a) Video frame interpolation task (VFI). (b) RS correction task using neighboring frames (RSC). (c) RS correction task using dual frames with reversed RS distortion (Dual-RSC). (d) Blurry frame interpolation task (BFI). (e) Joint RS correction and interpolation task using neighboring frames (RSCI). (f) Joint RS correction and interpolation task using dual frames with reversed RS distortion (Dual-RSCI).
	}
	\label{fig:tasks}
\end{figure*}

Fan and Dai~\cite{fan2021inverting} proposed a Rolling Shutter temporal Super-Resolution (RSSR) pipeline for this joint interpolation and correction task. Under the assumption of constant velocity of camera motion and a static scene, RSSR combines a neural network and a manual conversion scheme to estimate undistortion flow for a specific time instance based on the temporal correlation of two adjacent frames (See Fig.~\ref{fig:rs2mgs} for a variant using three consecutive frames). However, even without object motion, the undistortion flow learned in this way tends to overfit the training dataset, because of the intrinsic uncertainty of this setup especially the readout time for each row. As proved in~\cite{dai2016epipolar}, the relative motion of two adjacent RS frames is characterized by the generalized epipolar geometry, which requires at least 17 point matches to determine camera motion. Even worse, it suffers from non-trivial degeneracies, for example, when the camera translates along the baseline direction. In practice, both the relative motion velocity and readout setting will affect the magnitude of RS distortion, and the RSSR model and learning-based RS correction model~\cite{liu2020deep} tend to fail on samples with different readout setups, especially on real-world data with complex camera and/or object motion (See details in Sec.~\ref{sec:experiment} and the supplementary video).

To tackle this problem in dynamic scenes, modeling in the traditional way is particularly difficult, and the inconsistency in readout settings between training data and real test data is also challenging. Inspired by a novel dual-scanning setup~\cite{albl2020two} (bottom-to-top and top-to-bottom as shown in Fig.~\ref{fig:2rs2gs}) for rolling shutter correction, we argue that this dual setup is better constrained and bears more potential for dynamic scenes. Mathematically, it requires only 5 point matches to determine camera motion, which is much less than that required by the setup with two consecutive RS frames. The symmetric nature of the dual reversed distortion, \textit{i.e.} the start exposure times of the same row in two images are symmetric about the center scan line, implicitly preserves the appearance of the latent undistorted images. Thus, this setup can also help to bypass the effects of inconsistent readout settings. Regarding the hardware complexity and cost, we note that synchronized dual RS camera systems can be easily realized on multi-camera smartphones~\cite{albl2020two,yang2021real} and self-driving cars. Interpolation of dual RS images into GS image sequences provides a promising solution to provide robust RS distortion-free high-fps GS images instead of directly employing expensive high-fps GS cameras. This can be further served as a high-quality image source for high-level tasks such as SfM~\cite{zhuang2017rolling}, and 3D reconstruction~\cite{liu2020deep}.

Despite the strong geometric constraints arising from dual reversed distortion, it is still intractable to derive a video clip without prior knowledge from training data, as indicated in the large body of literature on video frame interpolation (VFI) from sparse GS frames (Fig. ~\ref{fig:gs2mgs}). Therefore, grounded upon the symmetric feature of the dual-RS setup, we design a novel end-to-end \textbf{I}ntermediate \textbf{F}rames \textbf{E}xtractor using \textbf{D}ual RS images with reversed distortion (IFED) to realize joint correction and interpolation. Inspired by~\cite{liu2020deep}, we introduce the dual RS time cube to allow our model to learn the velocity cube iteratively, instead of regressing directly to an optical flow cube, so as to promote convergence. A mask cube and residual cube learned from an encoder-decoder network are used to merge the results of two reversely distorted images after backward warping. Taking our result in Fig.~\ref{fig:teaser} as an example, the left image in the last row shows the mixed dual inputs $I^{(t)}_{t2b}$ (top-to-bottom scanning) and $I^{(t)}_{b2t}$ (bottom-to-top scanning) at time $t$. The rest of the row shows the extracted undistorted and smoothly moving frames by our method in chronological order.

To evaluate our method, we build a synthetic dataset with dual reversed distortion RS images and corresponding ground-truth sequences using high-fps videos from the publicly available dataset~\cite{nah2017deep} and self-collected videos. Besides, we also construct a real-world test set with dual reversed distortion inputs captured by a custom-made co-axial imaging system. Although similar concept of dual-RS~\cite{albl2020two} (stereo) setup and time field~\cite{liu2020deep} (2d) were proposed separately by previous works, we successfully combine and upgrade them to propose a simple yet robust architecture to solve the joint RS correction and interpolation (RS temporal super-resolution~\cite{fan2021inverting}) problem. The contributions of this work can be summarized as follows: 1) This is the first work that can extract video clips from distorted image in dynamic scenes. Besides, our solution can overcome the generalization problem caused by distinct readout settings. 2) We propose a novel end-to-end network architecture (IFED) that can iteratively estimate the accurate dual optical flow cube using pre-defined time cube and efficiently merges the symmetric information of the dual RS inputs for latent GS frame extraction. 3) Extensive experimental results demonstrate the superior accuracy and robustness of IFED against the state-of-the-art both on synthetic dataset and real-world data.

\section{Related Works}
In this section, we briefly review the closely related research on video frame interpolation and rolling shutter correction.

\subsection{Video Frame Interpolation}
Most existing solutions to VFI utilize optical flows to predict intermediate frames of captured images. These methods warp the input frames in a forward or backward manner based on the flow estimated by off-the-shelf networks, such as PWCNet~\cite{sun2018pwc}, FlowNet~\cite{dosovitskiy2015flownet,ilg2017flownet}, and RAFT~\cite{teed2020raft}. The warped frame is then refined by convolutional neural networks (CNNs) to obtain better visual quality. For example, SuperSlomo~\cite{jiang2018super} uses a linear combination of two bi-directional flows from an off-the-shelf network for intermediate flow estimation and performs backward warping to infer latent frames. DAIN~\cite{bao2019depth} further improves the intermediate flow estimation by employing a depth-aware flow projection layer. Recently, RIFE~\cite{huang2020rife} achieves high-quality and real-time frame interpolation with an efficient flow network and a leakage distillation loss for direct flow estimation. In contrast to backward warping, \el{Niklaus}~\cite{niklaus2020softmax} focuses on forward warping interpolation by proposing Softmax splatting to address the conflict of pixels mapped to the same target location. On the other hand, some recent works~\cite{choi2020channel,kalluri2020flavr} achieve good results using flow-free methods. For example, CAIN~\cite{choi2020channel} employs the PixelShuffle operation with channel attention to replace the flow computation module, while FLAVR~\cite{kalluri2020flavr} utilizes 3D space-time convolutions instead to improve efficiency and performance on non-linear motion and complex occlusions. 

VFI includes a branch task, called blurry frame interpolation~\cite{jin2018learning,purohit2019bringing,jin2019learning,shen2020blurry}, which is analogous to our target problem. In this task, a blurry image is a temporal average of sharp frames at multiple instances. The goal is to deblur the video frame and conduct interpolation, as illustrated in Fig.~\ref{fig:bgs2mgs}. \el{Jin}~\cite{jin2018learning} proposed a deep learning scheme to extract a video clip from a single motion-blurred image. For a better temporal smoothness in the output high-frame-rate video, \el{Jin}~\cite{jin2019learning} further proposed a two-step scheme consisting of a deblurring network and an interpolation network. Instead of using a pre-deblurring procedure, BIN~\cite{shen2020blurry} presents a multi-scale pyramid and recurrent architecture to reduce motion blur and upsample the frame rate simultaneously. Other works~\cite{pan2019bringing,lin2020learning} utilize additional information from event cameras to bring a blurry frame alive with a high frame rate.

Existing VFI methods ignore the distortions in videos captured by RS cameras. In our work, instead of considering RS distortion as a nuisance, we leverage the information embedded in it to retrieve a sequence of GS frames.

\subsection{Rolling Shutter Correction}

RS correction itself is also a highly ill-posed and challenging problem. Classical approaches~\cite{forssen2010rectifying,baker2010removing,oth2013rolling} work under some assumptions, such as a static scene and restricted camera motion (\textit{e.g.}, pure rotations and in-plane translations). Consecutive frames are commonly used as inputs to estimate camera motion for distortion correction. \el{Grundmann}~\cite{grundmann2012calibration} models the motion between two neighboring frames as a mixture of homography matrices. \el{Zhuang}~\cite{zhuang2017rolling} develops a modified differential SfM algorithm for estimating the relative pose between consecutive RS frames, which in turn recovers a dense depth map for RS-aware warping image rectification. \el{Vasu}~\cite{vasu2018occlusion} sequentially estimates both camera motion and the structure of the 3D scene that accounts for the RS distortion, and then infers the latent image by performing depth and occlusion-aware rectification. \el{Rengarajan}~\cite{rengarajan2016bows} corrects the RS image according to the rule of ``straight-lines-must-remain-straight''. \el{Purkait}\cite{purkait2017rolling} assumes that the captured 3D scene obeys the Manhattan world assumption and corrects the distortion by jointly aligning vanishing directions.

In recent years, learning-based approaches have been proposed to address RS correction in more complex cases. \el{Rengarajan}~\cite{rengarajan2017unrolling} builds a CNN architecture with long rectangular convolutional kernels to estimate the camera motion from a single image for RS correction. \el{Zhuang}~\cite{zhuang2019learning} uses two independent networks to predict a dense depth map and camera motion from a single RS image, implementing RS correction as post-processing. \el{Liu}~\cite{liu2020deep} proposes a DeepUnrollNet to realize end-to-end RS correction with a differentiable forward warping block. SUNet~\cite{fan2021sunet} utilizes a symmetric consistency constraint of two consecutive frames to achieve state-of-the-art performance.

The most relevant research to ours are~\cite{fan2021inverting},~\cite{albl2020two} and the previously mentioned~\cite{liu2020deep}. \cite{fan2021inverting} proposed the first learning-based solution (RSSR) for latent GS video extraction from two consecutive RS images. On the other hand,~\cite{albl2020two} proposed a stereo dual-RS setup for RS correction task that infers an undistorted GS frame based on the geometric constraints among dual RS reversely distorted images. However, to the best of our knowledge, there are no methods able to achieve RS temporal super-resolution in dynamic scenes. Geometric constraints of ~\cite{fan2021inverting} and~\cite{albl2020two} are limited to static scenes. Besides, current learning-based methods including~\cite{liu2020deep,fan2021inverting} suffer from the inherent ambiguity of consecutive setup. We discover the merit of dual-RS to overcome distinct readout setups, which is not mentioned in~\cite{albl2020two}, and we upgrade the velocity field from~\cite{liu2020deep} to first time realize RS temporal SR in dynamic scenes.

\begin{figure}[!t]
\centering
\begin{minipage}[h]{0.52\textwidth}
\centering
\includegraphics[width=.95\textwidth]{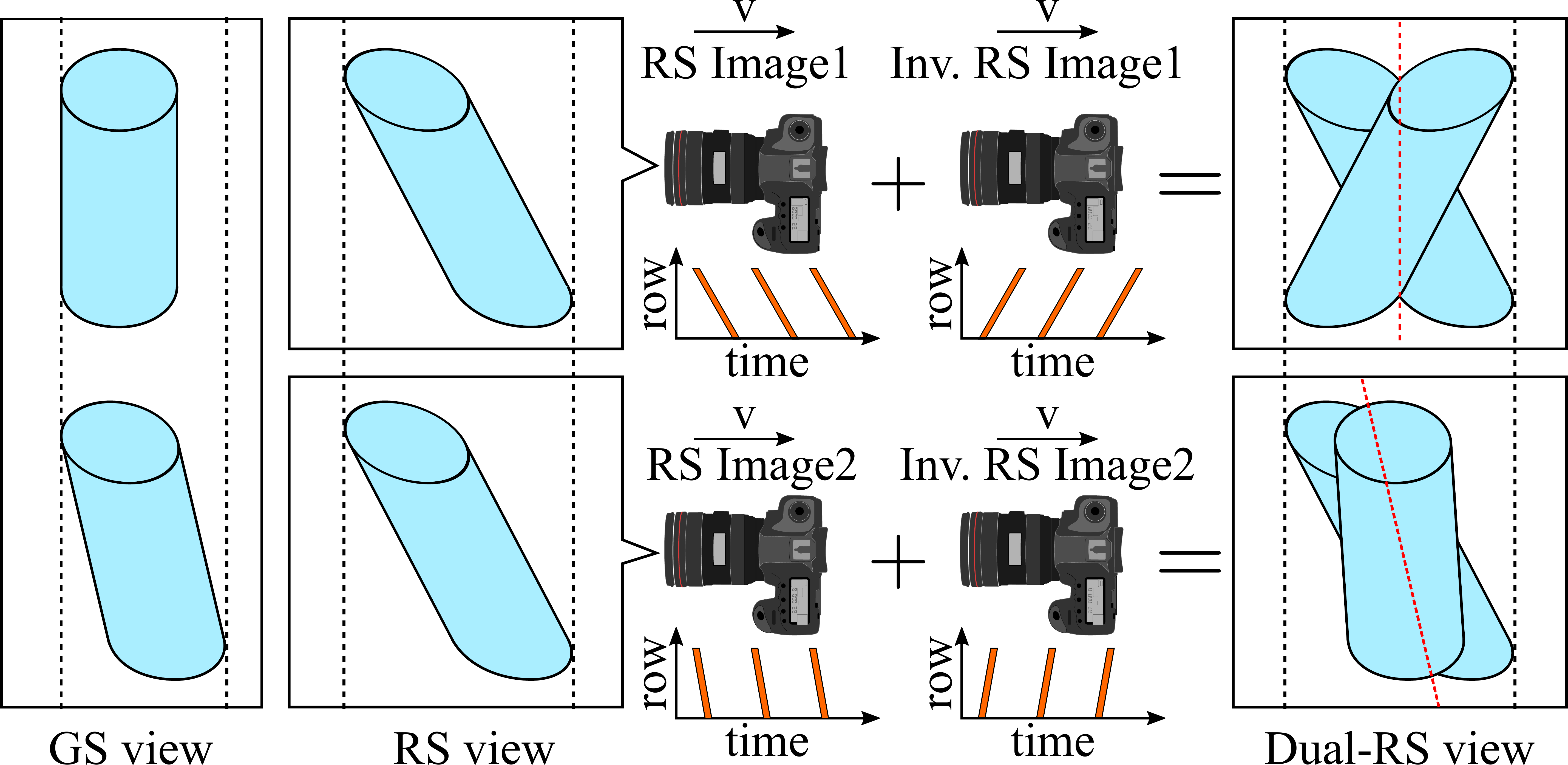}
\caption{\textbf{RS correction ambiguity.}}
\label{fig:ambiguity}
\end{minipage}
\hfill
\begin{minipage}[h]{0.46\textwidth}
\captionof{table}{\textbf{Details of RS-GOPRO}.}
\label{tab:rs-gopro_config}
\resizebox{\linewidth}{!}{
\begin{tabular}{lccc}
	\toprule
	&  train & validation & test\\
	\midrule
	sequences& 50 & 13 & 13\\
	RS images & 3554 ($\times$2) & 945 ($\times$2) & 966 ($\times$2) \\
	GS images & 31986 & 8505 & 8694 \\
	resolution& \multicolumn{3}{c}{960$\times$540}\\
	row exposure& \multicolumn{3}{c}{\SI{1.0}{ms}}\\
	row readout& \multicolumn{3}{c}{\SI{87}{\mu s}}\\
	\bottomrule
\end{tabular}
}
\end{minipage}
\end{figure}

\section{Joint RS Correction and Interpolation}
\label{sec:description}
In this section, we first formulate the joint RS correction and interpolation problem. Then, we introduce the datasets for validation and comparison.

\subsection{Problem Formulation}
\label{sec:problem}
An RS camera encodes temporal visual information in an image similar to a high-frame-rate GS camera that samples the scene rapidly but only takes one row of the scene each time. In our case, we do not consider the presence of blur. Formally, given an RS video ($\{ I_r^{(t)} \}$) and a GS video ($\{ I_g^{(t)} \}$), we can express each row ($i$) in an RS image ($I_r^{(t)}[i]$) in terms of its corresponding GS image ($I_g^{(t)}[i]$) through the following equation:
\begin{equation}
	I_r^{(t)}[i] = I_g^{(t+(i-M/2)t_r)}[i],
\end{equation}
where $t_r$ denotes the readout time for each RS row; $M$ denotes the total number of rows in the image; $t+(i-M/2)t_r$ is the time instant of scanning the $i^{th}$ row; and $I_g^{(t+(i-M/2)t_r)}[i]$ is the portion of the GS image that will appear in the RS image. Note that we define the time $t$ of an RS image $I_r^{(t)}$ as the midpoint of its exposure period (\textit{i.e.}, each RS image is captured from $t_s$ to $t_e$, where $t_s = t- t_rM/2$ and $t_e = t+ t_rM/2$).

The objective of the joint RS correction and interpolation is to extract a sequence of undistorted GS images ($\left\{ I_g^{(t)}, t\in \left[ t_s, t_e \right] \right\}$) from the RS images. Directly feeding an RS image ($I_r^{(t)}$) into a network $\mathcal{F}\left(I_r^{(t)}; \Theta \right)$, parameterized by the weight $\Theta$, to extract a sequence of GS images is infeasible without strong restrictions such as static scenes and known camera motions. A straightforward approach is to use temporal constraints from neighboring frames, such that the input is a concatenation of neighboring frames as $I_{inp}^{(t)} = \left\{ I_r^{(t-1/f)}, I_r^{(t)} \right\}$, where $f$ denotes the video frame rate. This is the case of RSSR~\cite{fan2021inverting}, which can easily overfit the readout setting of the training data. Theoretically, the generic RSC problem cannot be solved by using only consecutive frames. We show a common ambiguity of consecutive frames setup, using a toy example in Fig.~\ref{fig:ambiguity}. Suppose there are two similar cylinders, one of them is tilted, as shown in GS view. Then, two RS cameras moving horizontally at the same speed $v$ but with different readout time setups can produce the same RS view, \textit{i.e.}, a short readout time RS camera for the tilted cylinder and a long readout time RS camera for the vertical cylinder. Therefore, the models based on consecutive frames are biased to the training dataset. Although these models can correct RS images, they do not know how much correction is correct facing data beyond the dataset. Instead, we introduce another constraint setting that utilizes intra-frame spatial constraints of dual images taken simultaneously but with reversed distortion captured by top-to-bottom (t2b) and bottom-to-top (b2t) scanning. Formally, the optimization process is described as:
\begin{equation}\label{eq:solution}
	\widehat{\Theta} = \argmin_{\Theta} \left| \left\{ I_g^{(t)}, t\in \left[ t_s, t_e \right] \right\} - \mathcal{F}\left(I_{t2b}^{(t)}, I_{b2t}^{(t)}  ; \Theta\right) \right|,
\end{equation}
where $\widehat{\Theta}$ are optimized parameters for the joint task. $I_{t2b}^{(t)}$ denotes the t2b RS frame at time $t$, while $I_{b2t}^{(t)}$ denotes the b2t RS frame at the same time. We find that the dual-RS setup can avoid ambiguity because the correct correction pose can be estimated based on the symmetry, as shown in the dual-RS view. 

\subsection{Evaluation Datasets}
\label{sec:dataset}
\noindent \textbf{Synthetic Dataset.} For the pure RS correction task, the Fastec-RS~\cite{liu2020deep} dataset uses a camera mounted on a ground vehicle to capture high-fps videos with only horizontal motion. Then, RS images are synthesized by sequentially copying a row of pixels from consecutive high-fps GS frames. We synthesized a dataset for the joint RS correction and interpolation task in a similar way, but with more motion patterns and multiple ground truths for one input. High-fps GS cameras with sufficient frame rate to synthesize RS-GS pairs are expensive and cumbersome to operate. Thus, we chose a GoPro (a specialized sports camera) as a trade-off. Empirically, the GoPro's tiny RS effect causes negligible impact on the learning process of our task. Specifically, we utilize the high-fps (\SI{240}{fps}) videos from the publicly available GOPRO~\cite{nah2017deep} dataset and self-collected videos using a GoPro HERO9 to synthesize the dataset, which we refer to as RS-GOPRO. We first interpolated the original GS videos to \SI{15360}{fps} by using an off-the-shelf VFI method (RIFE~\cite{huang2020rife}), and then followed the pipeline of~\cite{liu2020deep} to synthesize RS videos. RS-GOPRO includes more complex urban scenes (\textit{e.g.}, streets and building interiors) and more motion patterns, including object-only motion, camera-only motion, and joint motion. We created train/validation/test sets (50, 13, and 13 sequences) by randomly splitting the videos while avoiding images of a video from being assigned into different sets. Regarding input and target pairs, there are two kinds of input RS images which have reversed distortion, and nine consecutive GS frames are taken as ground truth for the extracted frame sequence. The image resolution is 960$\times$540. The readout time for each row is fixed as \SI{87}{\micro\second}. Please see the details of RS-GOPRO in Table~\ref{tab:rs-gopro_config}.

\noindent \textbf{Real-world Test Set.} Inspired by~\cite{zhong2020efficient} and~\cite{albl2020two}, we built a dual-RS image acquisition system using a beam-splitter and two RS cameras that are upside down from each other to collect real-world data for validation. The readout setting of the proposed dual-RS system can be changed by replacing the type of RS camera (\textit{e.g.}, FL3-U3-13S2C, BFS-U3-63S4C). Please see details of our acquisition system in supplementary materials. We collect samples of various motion patterns, such as camera-only motion, object-only motion like moving cars and a rotating fan, and mixed motion. Each sample includes two RS distorted images with reversed distortion but without a corresponding ground truth sequence.

\begin{figure*}[!t]
  \centering
  \includegraphics[width=\linewidth]{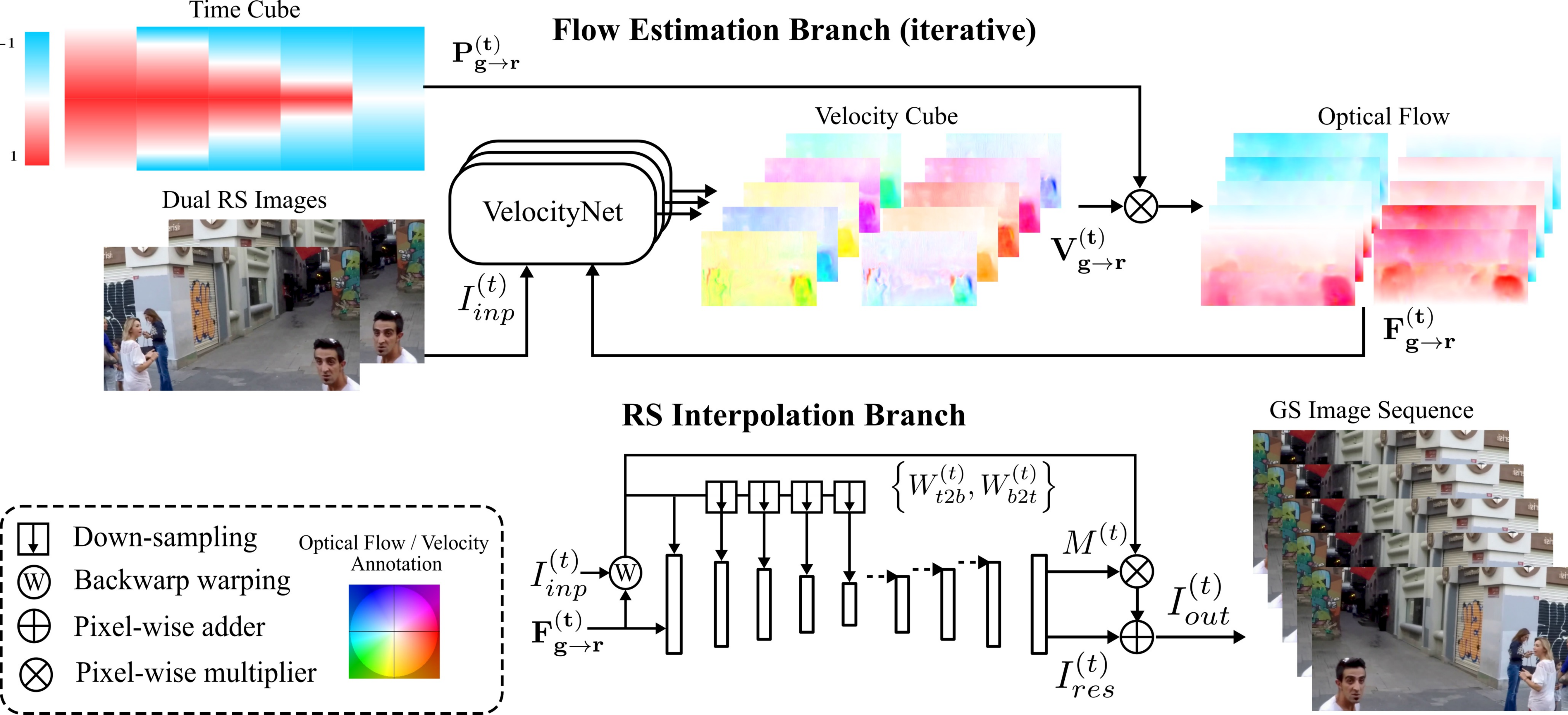}
  \caption{\textbf{Network architecture of IFED.} Note that the color annotation of the dual RS time cube is different from optical flow and velocity. It represents the relative time gap between each row and the time instance of the latent GS image. This architecture first utilizes dual RS along with time cube to iteratively estimate velocity cube for better optical flow learning. Then, the warped dual frames are combined together as complementary information through the mask and residual cube learned from the encoder-decoder network to make inferences for the underlying GS image sequences.
  }
  \label{fig:architectures}
\end{figure*}

\section{Methodology}
\label{sec:method}
We present the proposed architecture and implementation details in this section.

\subsection{Pipeline of IFED}
The proposed IFED model utilizes an architecture inherited from existing successful VFI methods~\cite{jiang2018super,huang2020rife}, including a branch to estimate the optical flow for backward warping and an encoder-decoder branch to refine the output (see Fig.~\ref{fig:architectures}). However, directly estimating optical flow from the latent GS image to the input RS image is challenging due to the intra-frame temporal inconsistency of an RS image. The optical flow from GS to RS is dependent on two variables: the time difference and relative velocity of motion. As we already know the scanning mechanism of the RS camera, we are able to obtain the time difference between the input RS image and the target GS image. Thus, we propose a dual time cube as an RS prior to decouple this problem, and let the model regress the dual velocity cube to indirectly estimate the corresponding dual optical flow cube. The number of time instances per dual cube is twice the number of extracted GS frames. These time instances are sampled uniformly from the entire RS exposure time (\textit{e.g.}, Fig.~\ref{fig:architectures} shows the extraction of 5 GS frames). There is an implicit assumption that the velocity field of each row of the extracted frame is constant. Considering the short exposure time of the actual RS image and the short percentage of time corresponding to the extracted GS frames, this assumption can be basically satisfied in most scenarios. Besides, the dual warped features can be further adjusted and merged by the interpolation branch, which enables our method to handle the challenging cases of the spinning fans and wheels with row-wise non-uniform velocity.

Specifically, assuming our target latent sequence has $N$ images, the target optical flow cube for one RS image can be expressed as follows:
\begin{equation}
	\mathbf{F_{g\to r}^{(t)}} = \left\{ F_{g\to r}^{(t_n)} \right\}, n\in \left\{1, \cdots, N\right\},
\end{equation}
where $t_n = t - t_rM\left(\frac{1}{2} - \frac{n}{N}\right)$ and $F_{g\to r}^{(t_n)}$ denotes the optical flow from the GS image at time $t_n$ to the distorted RS input $I_{r}^{(t)}$. Regarding the time cube, the values at row $m$ of the time map $P_{g\to r}^{(t)}$ are given by:
\begin{equation}
    P_{g\to r}^{\left(t_n\right)}[m] = \frac{m-1}{M-1} - \frac{n-1}{N-1}, m\in[1..M], n\in[1..N].
\end{equation}
Then, the RS time cube $\mathbf{P_{g\to r}^{(t)}} = \left\{ P_{g\to r}^{(t_n)} \right\}$ can be expressed in the same format as $\mathbf{F_{g\to r}^{(t)}}$. To obtain the optical flow cube, we need the network to generate a velocity cube $\mathbf{V_{g\to r}^{(t)}} = \left\{ V_{g\to r}^{(t_n)} \right\}$ and multiply it with the RS time cube as follows:
\begin{equation}
	\left\{ F_{g\to r}^{(t_n)} \right\} = \left\{ P_{g\to r}^{(t_n)}V_{g\to r}^{(t_n)} \right\}.
\end{equation}
Our flow branch uses several independent subnetworks (VelocityNet) to iteratively take dual RS images $I_{inp}^{(t)}$ and previously estimated dual optical flow cube as inputs for dual velocity cube $\mathbf{V_{g\to t2b}^{(t)}}$ estimation. The input scale (resolution) of the subnetwork are scaled sequentially in an iterative order following a coarse-to-fine manner (adjusted by bilinear interpolation). These sub-networks share the same structure, starting with a warping of the inputs, followed by a series of 2d convolutional layers. The initial scale velocity cube estimation is realized without the estimated optical flow cube. This branch is shown in the upper part of Fig.~\ref{fig:architectures}.

After obtaining the optical flow cube, we can generate a series of warped features and the warped dual RS images $W^{(t)} = \left\{W_{b2t}^{(t)}, W_{t2b}^{(t)}\right\}$ as multi-scale inputs to an encoder-encoder network with skip connections for merging results. Specifically, a residual cube $I_{res}^{(t)}$ and a dual mask cube $M^{(t)}$ are generated to produce the final frame sequence (See the bottom part of Fig.~\ref{fig:architectures}) as follows:
\begin{equation}
	I_{out}^{(t)} = I_{res}^{(t)} + M^{(t)}W_{t2b}^{(t)} + \left(1 - M^{(t)}\right)W_{b2t}^{(t)}.
\end{equation}

\subsection{Implementation Details}
We implement the method using PyTorch~\cite{paszke2019pytorch}. There are three 4 sub-networks in the flow network branch for velocity cube learning, each with eight $3\times 3$ convolutional layers. The inputs scale is gradually adjusted from $1/8$ to original size as the channel size is reduced. The network is trained in 500 epochs. The batch size and learning rate are equal to 8 and $1\times10^{-4}$ separately. AdamW~\cite{loshchilov2017decoupled} is used to optimize the weights with a cosine annealing scheduler. The learning rate is gradually reduced to $1\times10^{-8}$ throughout the whole process. $256\times256$ cropping is applied for both dual RS images and the time cube. Because the relative time difference between the same row of adjacent crops is constant, training with cropping does not affect the full frame inference. More details of the sub-networks and cropping are in supplementary materials. The loss function to train the model is given by:
\begin{equation}
	\mathcal{L} = \mathcal{L}_{char} + \lambda_p\mathcal{L}_{perc} + \lambda_v\mathcal{L}_{var},
\end{equation}
where $\mathcal{L}_{char}$ and $\mathcal{L}_{perc}$ denote the Charbonnier loss and perceptual loss~\cite{johnson2016perceptual} for the extracted frame sequence; while $\mathcal{L}_{var}$ denotes the total variation loss for the estimated flows, to smooth the warping. $\lambda_p$ and $\lambda_v$ are both set to 0.1.

\begin{figure*}[!t]
	\centering
	\includegraphics[width=\textwidth]{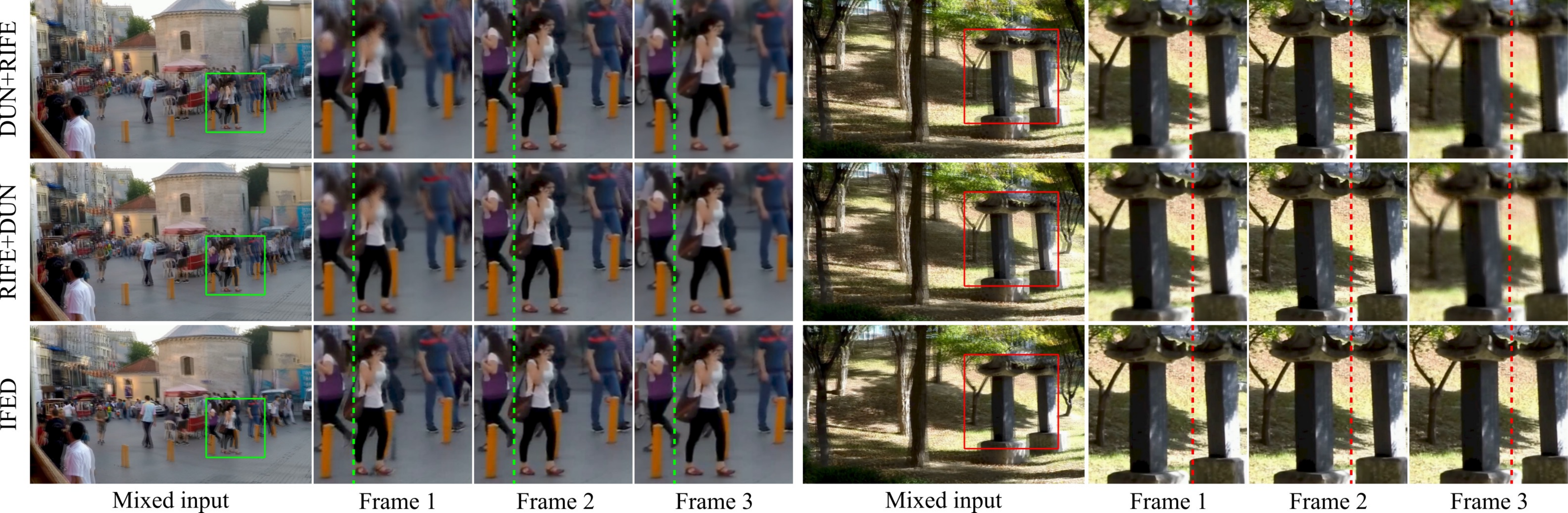}
	\caption{\textbf{Visual results on RS-GOPRO.} Zoom-in results are shown chronologically on the right side of the mixed input. IFED restores the smooth moving sequence with clearer details while cascaded scheme introduced unclear artifacts.}
	\label{fig:comparison_rs-gopro}
\end{figure*}

\begin{figure}[!t]
	\centering
	\includegraphics[width=\textwidth]{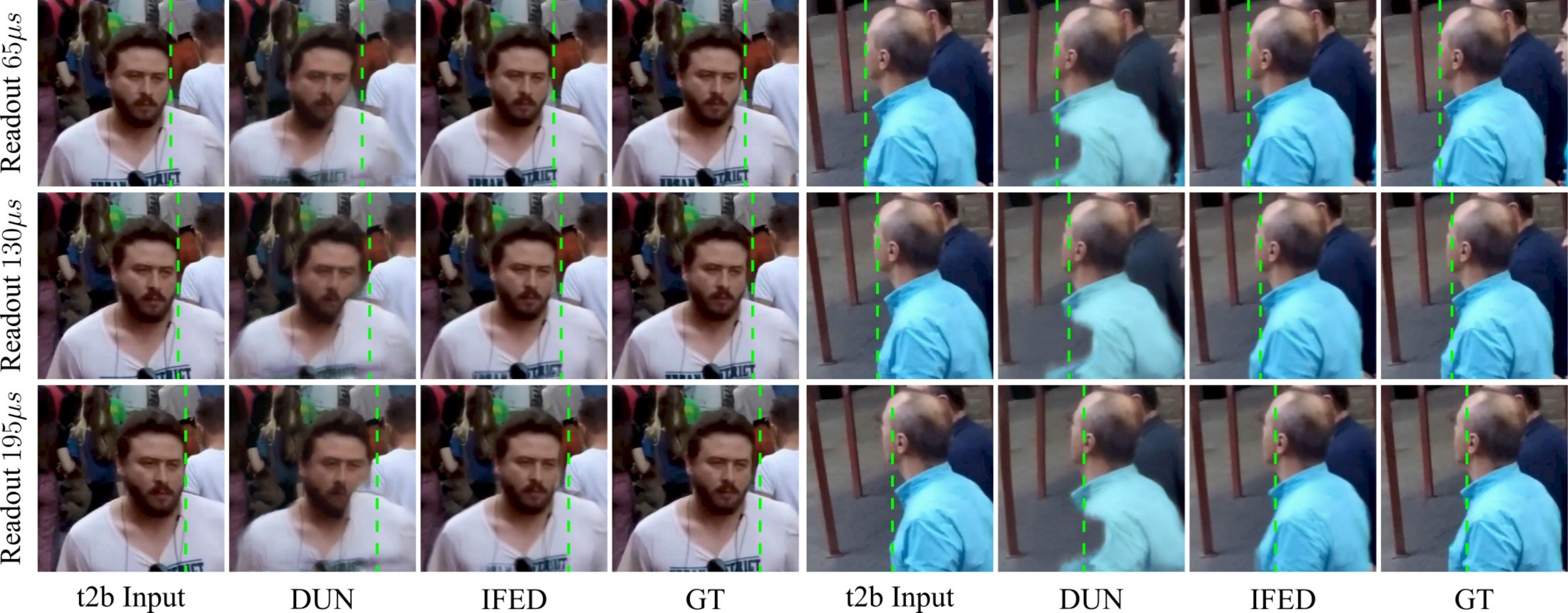}
	\caption{\textbf{Generalization ability on distinct readout time settings.} Both our IFED and DUN~\cite{liu2020deep} are trained on fixed readout setting, while IFED can successfully generalize to different readout settings from \SI{65}{\mu s} to \SI{195}{\mu s}.
	}
	\label{fig:readout_comparison}
\end{figure}

\section{Experimental Results}
\label{sec:experiment}
In this section, we first present comparison experiments on the synthesized dataset RS-GOPRO in Sec.~\ref{sec:synthetic_data}. Next, we show the generality of our method on real-world data in Sec.~\ref{sec:real_data}. Finally, we present the ablation study in Sec.~\ref{sec:ablation}. Please see more additional experimental results in our appendix.

\subsection{Results on Synthetic Dataset}
\label{sec:synthetic_data}
\begin{figure}[!t]
\centering
\begin{minipage}[h]{0.48\textwidth}
\centering
\captionof{table}{\textbf{Quantitative results on RS-GOPRO.} f\# denotes \# of frames extracted from the input RS images.}
\label{tab:rs-gopro_comparison}
\resizebox{\linewidth}{!}{
\begin{tabular}{lccc}
	\toprule
	& PSNR $\uparrow$ & SSIM $\uparrow$ & LPIPS $\downarrow$\\
	\midrule
	\noalign{\smallskip}
	DUN (f1) & 26.37 & 0.836 & 0.058\\
	DUN + RIFE (f3) & 25.38 & 0.788 & 0.159 \\
	DUN + RIFE (f5) & 25.45 & 0.798 & 0.111 \\
	DUN + RIFE (f9) & 25.31 & 0.795 & 0.102 \\
	RIFE + DUN (f3) & 23.05 & 0.719 & 0.124 \\
	RIFE + DUN (f5) & 22.28 & 0.692 & 0.118 \\
	RIFE + DUN (f9) & 21.88 & 0.677 & 0.113 \\
	\midrule
	\noalign{\smallskip}
	IFED (f1) & 32.07 & 0.934 & 0.028 \\
	IFED (f3) & 28.48 & 0.872 & 0.058 \\
	IFED (f5) & 29.79 & 0.897 & 0.049 \\
	IFED (f9) & 30.34 & 0.910 & 0.046 \\
	\bottomrule
\end{tabular}
}
\end{minipage}
\hfill
\begin{minipage}[h]{0.46\textwidth}
\centering
  \includegraphics[width=\linewidth]{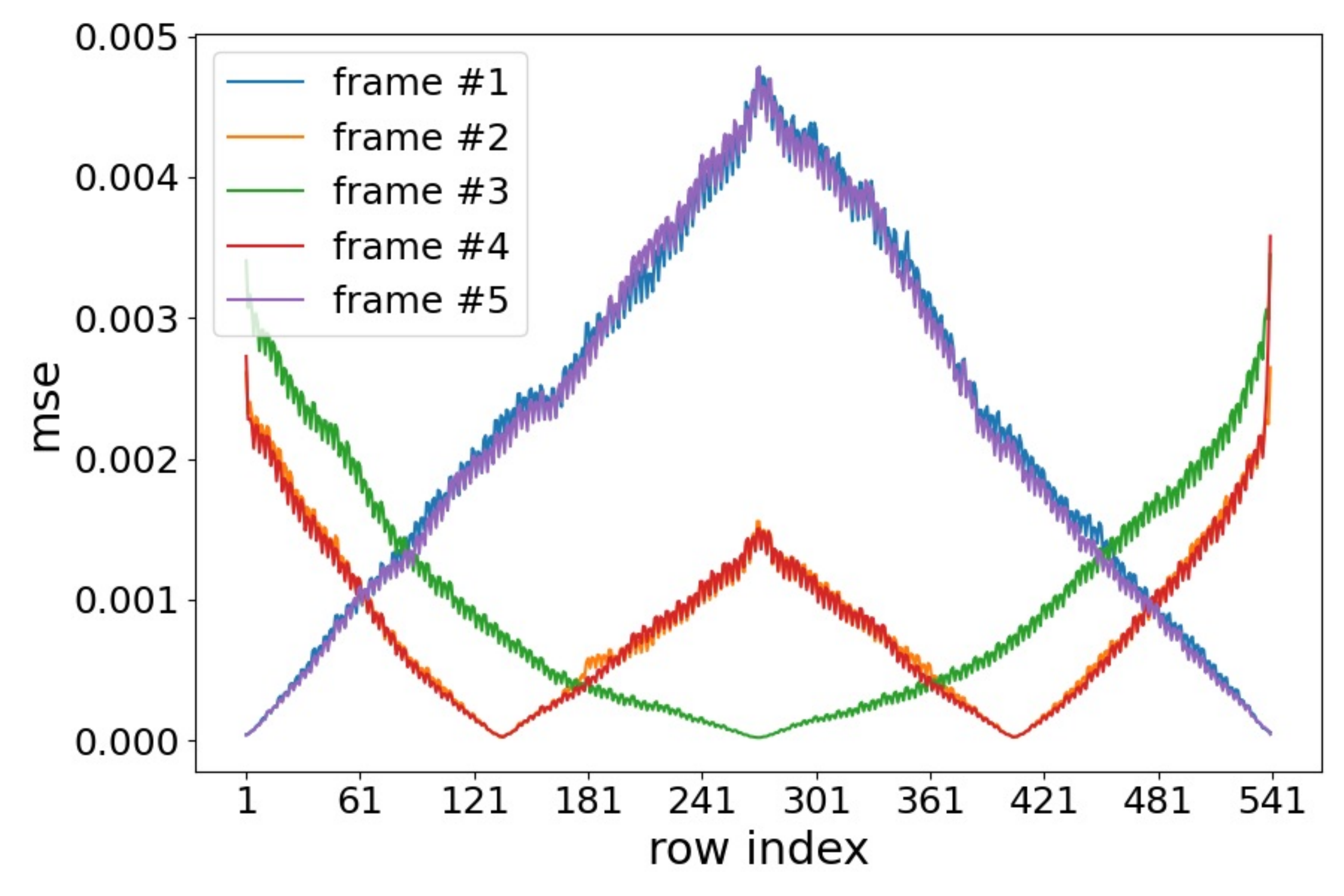}
  \figcaption{\textbf{Image mean squared errors based on row number in the case of IFED (f5).}}
  \label{fig:row_number}
\end{minipage}
\end{figure}

We implemented cascade schemes with RSC model DUN (DeepUnrollNet~\cite{liu2020deep}) and a VFI model RIFE~\cite{huang2020rife} using adjacent frames as inputs. Both orderings were examined, \textit{i.e.}, DUN+RIFE and RIFE+DUN ((b)+(a) and (a)+(b) in Fig.~\ref{fig:tasks}). We retrained DUN and RIFE on our dataset for extracting 1, 3, 5, and 9 frames for fair comparison. Quantitative results are shown in Table~\ref{tab:rs-gopro_comparison}. Over the different extracted frame settings, IFED shows superiority over the cascade schemes. The average performance of IFED is worst when the number of extracted frames is 3. Our interpretation is that the task degrades to a relatively easy RS correction task when the number of extracted frames is 1, while the greater continuity between extracted frames is better for convergence when the number of extracted frames is greater than 3. Qualitative results are shown in Fig.~\ref{fig:comparison_rs-gopro}. With the cascade schemes, the details are blurry, while ours are much clearer.

To verify the generalization on distinct readout settings, we synthesized RS images with distinct readout settings such as 65\si{\mu s}, 130\si{\mu s}, and 195\si{\mu s}. As illustrated in Fig.~\ref{fig:readout_comparison}, both our IFED and DUN~\cite{liu2020deep} are trained on fixed readout setting, while our IFED can successfully generalize to different readout settings without introducing artifacts and undesired distortions.

Besides, an row-wise image error analysis (f5) is shown in Fig.~\ref{fig:row_number} in terms of MSE. It indicates that the performance of a given row index depends on the minimum time (the smaller the better) between the row of that extracted GS frame and the corresponding rows of dual RS frames.

\subsection{Results on Real-world Data}
\label{sec:real_data}
\begin{figure*}[!t]
	\centering
	\includegraphics[width=\textwidth]{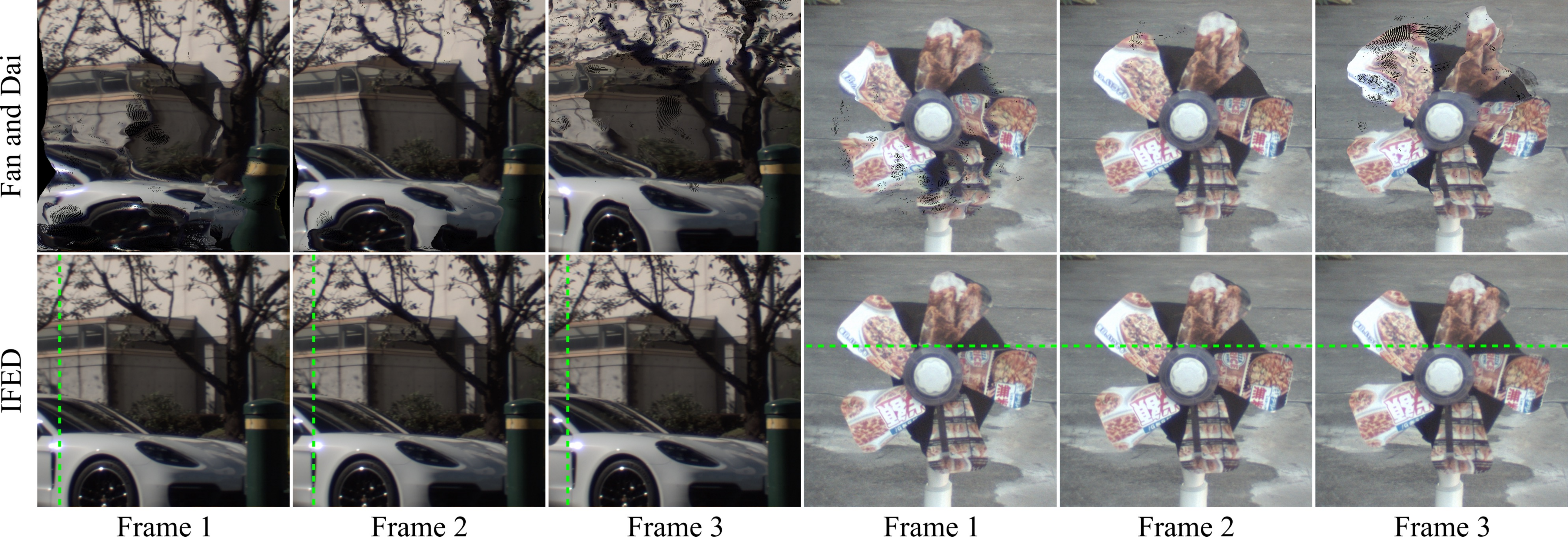}
	\caption{\textbf{Comparison with Fan and Dai~\cite{fan2021inverting} on real data.} Our results (the $2^{nd}$ row) are significantly better than Fan and Dai's for objects under both horizontal and rotational movements. Please refer to our supplementary videos.}
	\label{fig:vfi_real_data}
\end{figure*}

\begin{figure*}[!t]
	\centering
	\includegraphics[width=\textwidth]{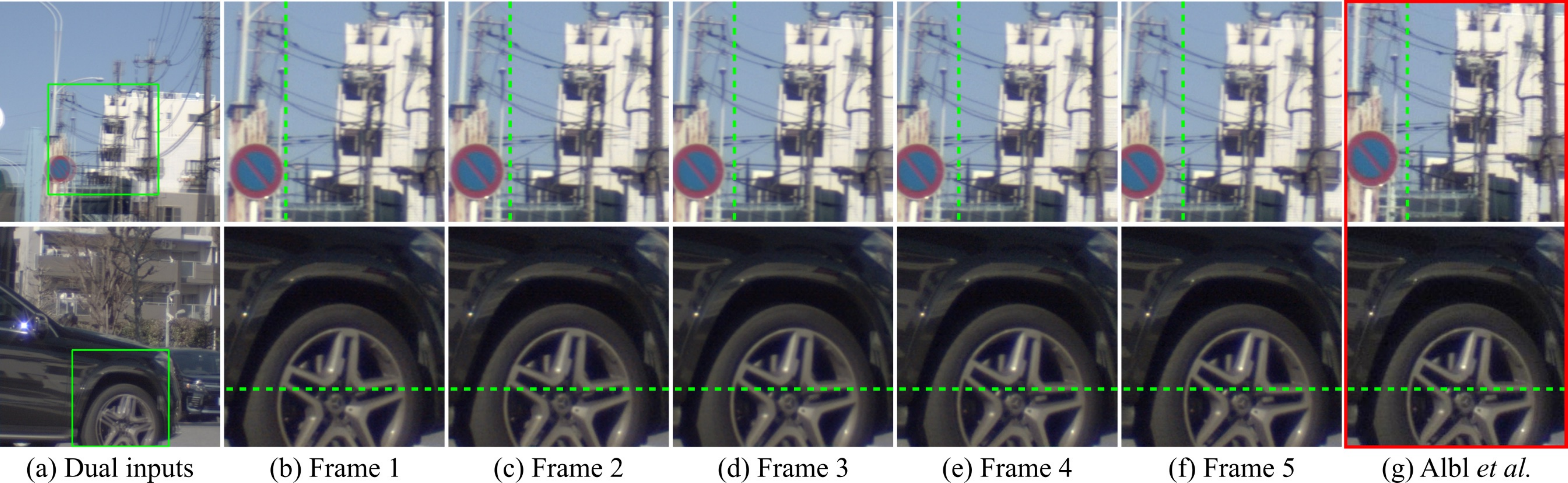}
	\caption{\textbf{Comparison with~\el{Albl}~\cite{albl2020two} on real data.} Both our method and \el{Albl}'s use the same dual inputs (the $1^{st}$ column). Our method brings the dual input alive by creating a sequence of images (Frame $1\sim5$), compared to one static image from \el{Albl}'s.}
	\label{fig:dual_real_data}
\end{figure*}

We also compare our method to the only existing work on extracting a GS sequence from RS images RSSR~\cite{fan2021inverting} and the only work for dual reversed RS image correction~\cite{albl2020two}. Since the source codes of these two works are not publicly available, we sent our real-world samples from different type of cameras to the authors for testing. The comparison results with RSSR~\cite{fan2021inverting} are shown in Fig.~\ref{fig:vfi_real_data}. RSSR cannot generalize to either the case of camera-only motion (the left example) or the case of object-only motion (the right example), while IFED is robust to different motion patterns. The visual results of IFED and~\cite{albl2020two} are illustrated in Fig.~\ref{fig:dual_real_data}. It demonstrates the ability of IFED to go beyond~\cite{albl2020two} by being able to extract a sequence of GS images in dynamic scenes, rather than just a single GS image in static scenes, from a dual reversed RS image. More results of IFED on the real dataset can be found in the supplementary materials.

\subsection{Ablation Study}
\label{sec:ablation}
\begin{figure}[!t]
	\centering
	\includegraphics[width=\textwidth]{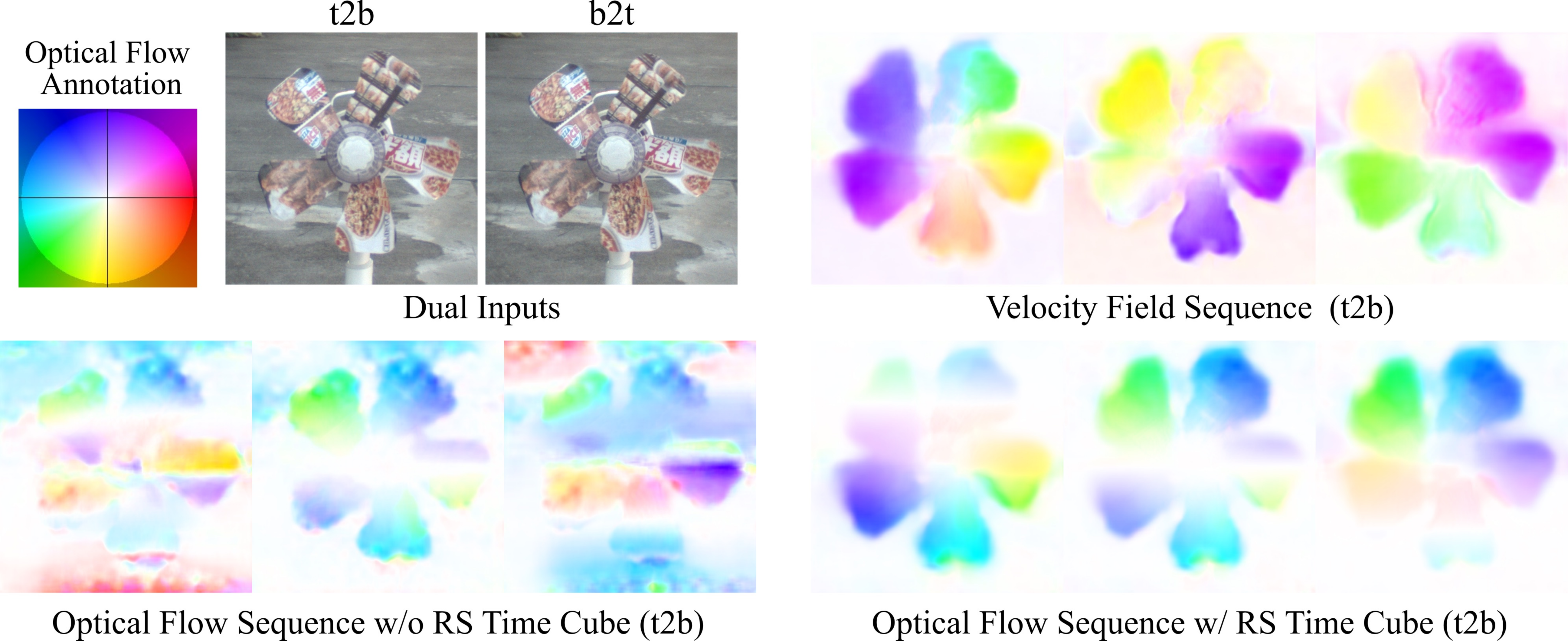}
	\caption{\textbf{Visualization of optical flow and velocity cube.} Equally with dual RS frames as input, using RS time cube prior to learn velocity cube can reduce the difficulty of optical flow learning and ultimately improve the flow quality.}
	\label{fig:flow}
\end{figure}

Table~\ref{tab:ablation} shows the results of our ablation study on the RS time cube prior. It shows that IFED without the prior generally leads to worse results, and the difference increases with a larger number of frames. Note that when the number of extracted frames equals 1, IFED \textit{w/o pr} can achieve better performance. The reason is that the task simply becomes the RSC task in this case, and the model can directly learn a precise flow for the middle time instance using dual RS inputs. When the number of extracted frames increases, the model needs the time cube to serve as an ``anchor'' for each time instance to improve the temporal consistency of the learned flow. We show visualizations of the flow and velocity cube with RS time cube prior and the flow cube without the prior in Fig.~\ref{fig:flow}. The flow sequence estimated without the RS time cube prior exhibits poor quality and consistency in time.

\begin{wraptable}{r}{5.5cm}
	\begin{center}
	\caption{\textbf{Ablation study for the prior.} \textit{w/o pr} denotes ``without RS time cube prior''.
	}
	\label{tab:ablation}
	\resizebox{\linewidth}{!}{
		\begin{tabular}{lccc}
			\toprule
			& PSNR $\uparrow$ & SSIM $\uparrow$ & LPIPS $\downarrow$\\
			\midrule
			& \multicolumn{3}{c}{Refer to IFED (f\#)} \\
			\cline{2-4}
			\noalign{\smallskip}
			IFED \textit{w/o pr} (f1) & +0.50 & +0.006 & -0.003 \\
			IFED \textit{w/o pr} (f3) & -0.40 & -0.008 & 0.000 \\
			IFED \textit{w/o pr} (f5) & -0.50 & -0.009 & +0.001 \\
			IFED \textit{w/o pr} (f9) & -0.70 & -0.012 & +0.001 \\
			\bottomrule
		\end{tabular}
	}
	\end{center}
\end{wraptable}

\section{Conclusions}
In this paper, we addressed a challenging task of restoring consecutive distortion-free frames from RS distorted images in dynamic scenes. We designed an end-to-end deep neural network IFED for the dual-RS setup, which has the advantages of being able to model dynamic scenes and not being affected by distinct readout times. The proposed dual RS time cube for velocity cube learning improves performance by avoiding direct flow estimation from the GS image to the RS image. Compared to the cascade scheme with existing VFI and RSC models as well as RSSR which takes temporally adjacent frames as inputs to do the same task, our IFED shows more impressive accuracy and robustness for both synthetic data and real-world data with different motion patterns.

\section*{Acknowledgement}
\label{sec:acknowledgement}
This work was supported by D-CORE Grant from Microsoft Research Asia, JSPS KAKENHI Grant Numbers 22H00529, and 20H05951, and JST, the establishment of university fellowships towards the creation of science technology innovation, Grant Number JPMJFS2108.

%
%
\bibliographystyle{splncs04}
\bibliography{egbib}

\newpage
\appendix
\section{More Details}
\subsection{Dual-RS Camera System}
\begin{figure}[!t]
	\centering
	\includegraphics[width=\textwidth]{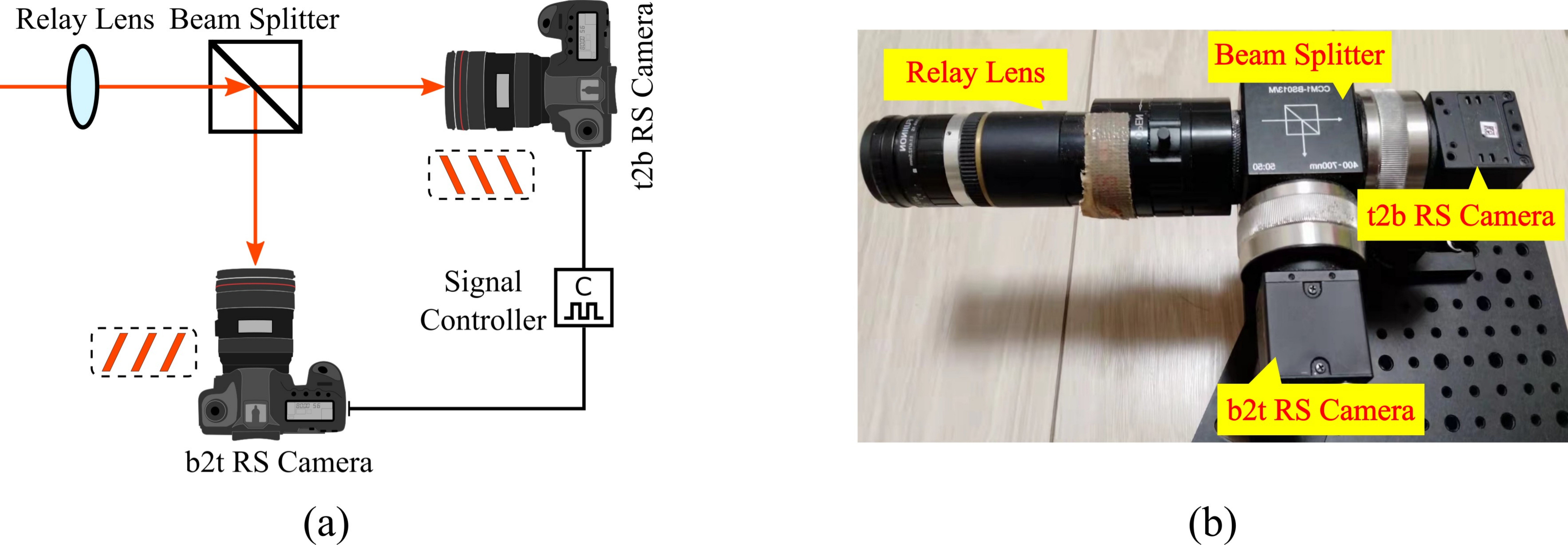}
	\caption{\textbf{Beam-splitter-based dual-RS acquisition system.} (a) is system schematic diagram. (b) shows real system used to collect dual-RS videos.}
	\label{fig:device}
\end{figure}

To evaluate our method and related works~\cite{fan2021inverting,albl2020two} on real-world data, we built a beam-splitter-based dual-RS camera system like~\cite{zhong2020efficient,rim2020real}, as illustrated in Fig.~\ref{fig:device}. One camera is installed upside down, such that the two RS cameras have reversed scanning directions. One system with two FL3-U3-13S2C rolling shutter cameras, and the other with two BFS-U3-63S4C rolling shutter cameras have been implemented, with different readout settings. The row-wise exposure time was properly adjusted to avoid motion blur.

\subsection{Structure of VelocityNet}
\begin{figure}[!t]
	\centering
	\includegraphics[width=.9\textwidth]{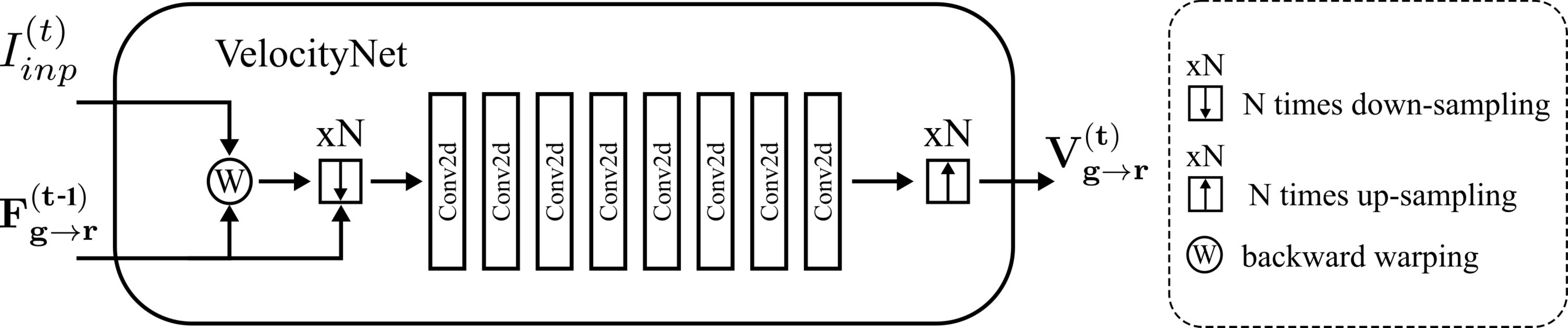}
	\caption{\textbf{Structure of VelocityNet for velocity cube estimation.}}
	\label{fig:velocitynet}
\end{figure}

The details of the subnetwork (VelocityNet) to estimate velocity cube is illustrated in Fig.~\ref{fig:velocitynet}. We totally use 4 subnetworks to iteratively take dual RS images $I_{inp}^{(t)}$ and previously estimated dual optical flow cube $F_{g\to r}^{(t-1)}$ as inputs for dual velocity cube $\mathbf{V_{g\to t2b}^{(t)}}$ estimation. The final velocity cube is equal to the sum of each subnetwork. These sub-networks share the same structure, starting with a warping of the inputs, followed by a series of 2d convolutional layers. All subnetworks have 8 convolutional layers. Before convolution, the scale (resolution) of the warped dual images and optical flow are scaled by linear interpolation. The scale ratio follows a coarse-to-fine manner from the first subnetwork to the last subnetwork, as $1/8$, $1/4$, $1/2$, and $1$, respectively. While the dimension of channel is set as 192, 128, 96 and 48, respectively. Note that the initial scale velocity cube estimation is realized without the estimated optical flow cube.

\subsection{Training with Cropping}
\begin{figure}[!t]
	\centering
	\includegraphics[width=.9\textwidth]{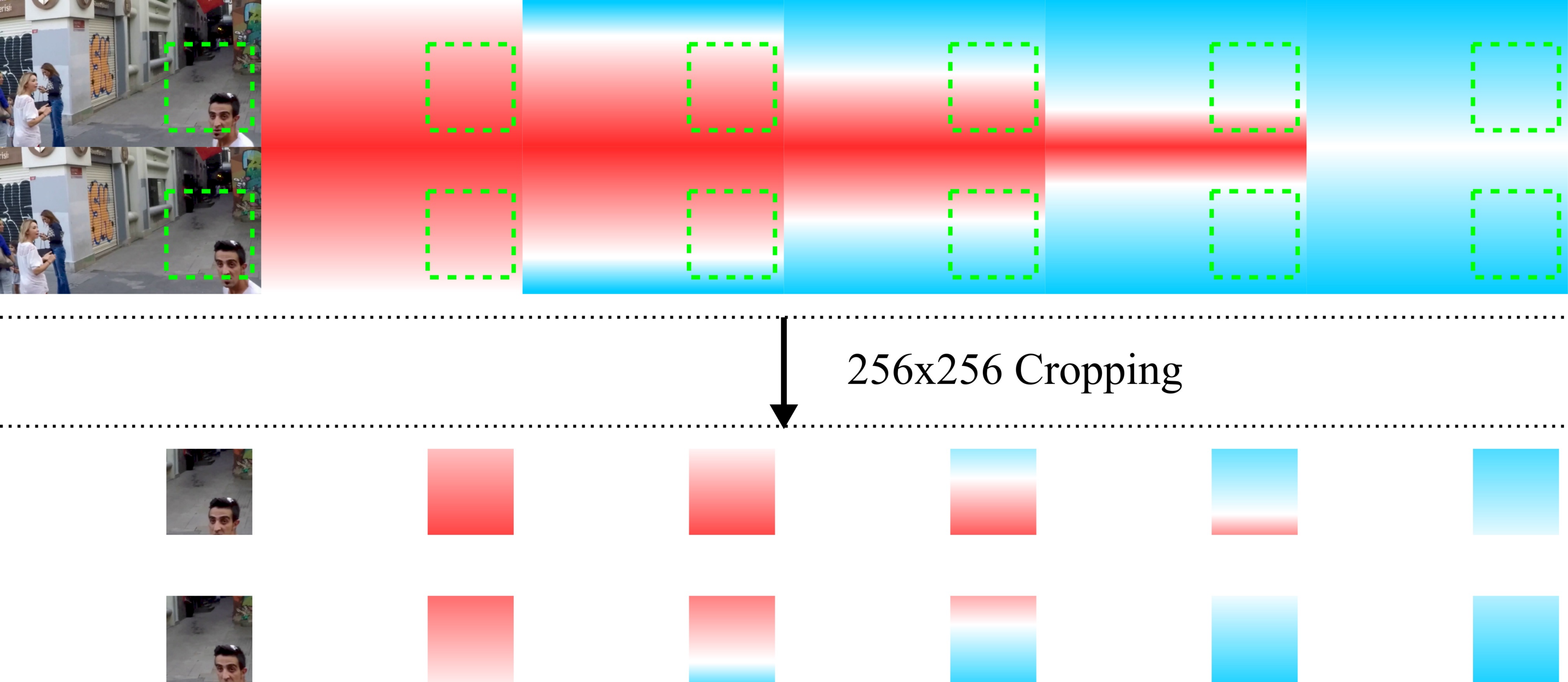}
	\caption{\textbf{Illustration of random cropping for training IFED (f5).}}
	\label{fig:cropping}
\end{figure}

We train the proposed IFED using $256\times 256$ random cropping for data augmentation. Taking the example of extracting 5 frames, the corresponding dual time cube prior will be cropped at the same position, as illustrated in Fig.~\ref{fig:cropping}.

\section{Additional Results}
\subsection{Video Results}
This work targets extracting undistorted image sequences from rolling shutter images with dual reversed distortion. The closest research to ours is Fan and Dai~\cite{fan2021inverting}, in which the RSSR algorithm was proposed to extract undistorted frames from two consecutive rolling shutter images. The authors did not release their source code of RSSR, yet kindly agreed to run a few real-world samples for fair comparison. For RSSR, we used two consecutive frames from one camera as input, while for our IFED, we used two dual reserved images (one from each camera) as input. Note that, although RSSR was not trained on our training set, we believe the comparison is fair and meaningful, in the sense that, both algorithms were trained on their own synthetic data, and both were tested on images captured by third-party cameras beyond the training set.

We present video results as \textcolor{red}{\texttt{results.mp4}}, including results of IFED on RS-GOPRO, as well as results of RSSR and IFED on real-world data. The video clips generated by IFED are more natural and visually appealing, while RSSR cannot generalize on real-world data. RSSR failed because their model only works when there are no moving objects in the scene, a restrictive assumption that rarely holds in practice. Also, the fact that the readout settings are not consistent between the training data and the real test data also poses challenges to RSSR.

\subsection{Comparison with Cascaded Schemes on Real-world Data}
\begin{figure}[!t]
	\centering
	\includegraphics[width=0.8\textwidth]{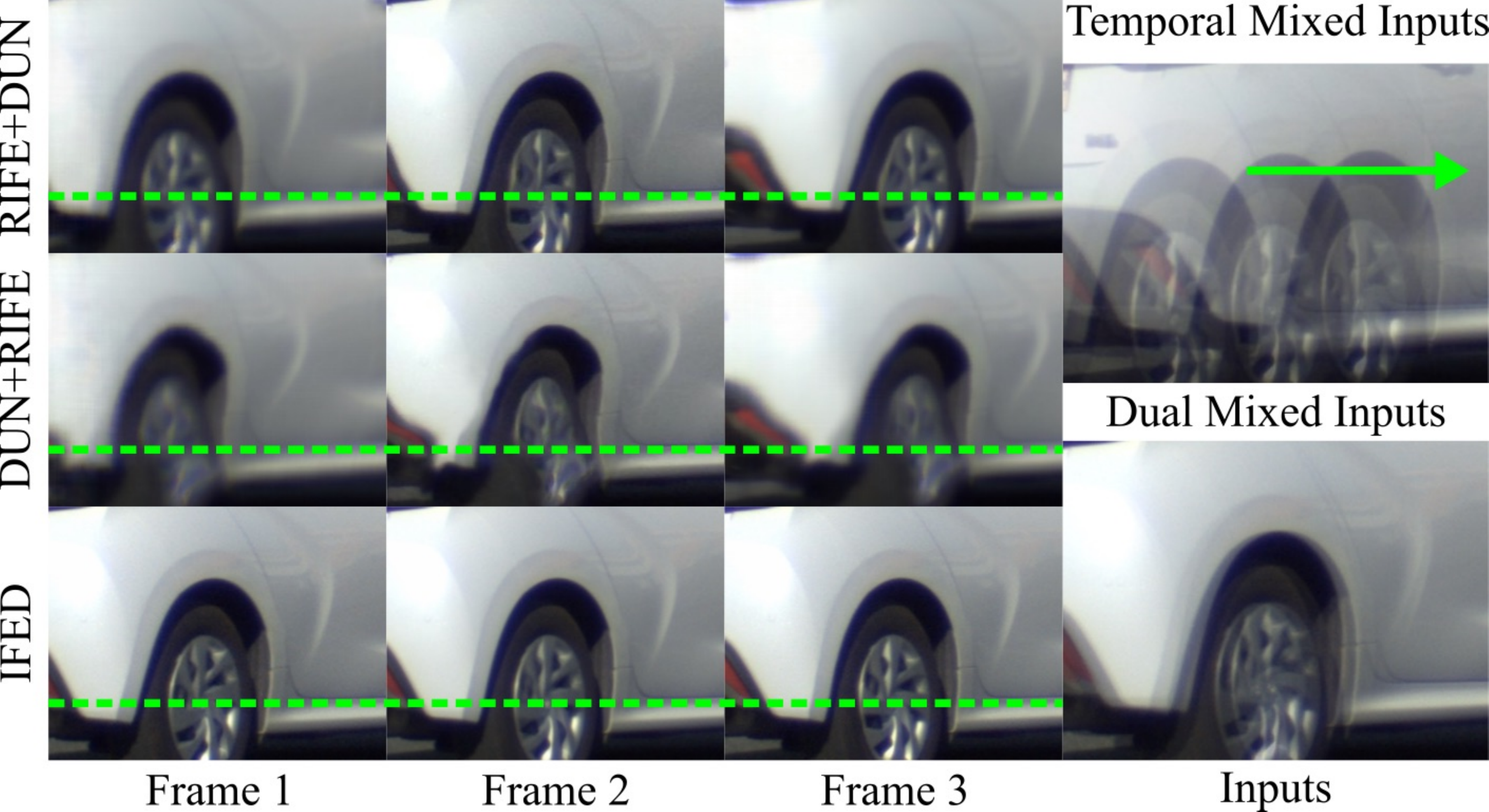}
	\caption{\textbf{Comparison with cascade schemes on real data.} Zoom-in results are shown on the left side of the input. Note that IFED uses the dual mixed inputs and the rest two methods use the temporal mixed inputs.}
	\label{fig:cascade_real_data}
\end{figure}

We show visual results for cascade schemes and IFED on real-world data in Fig.~\ref{fig:cascade_real_data} for the case of f3. The results are consistent with those on synthetic data where IFED produces sharper details than RIFE+DUN and DUN+RIFE. Note that IFED even correctly recovers the rotation of the wheel.

\subsection{The Effects of The Number Frames Extracted}
We show the cost of each IFED variant to infer a frame of size 960$\times$540 in Table~\ref{tab:inference_time}. When we implement models that extract different number of frames, we keep the parameters at almost the same level. The test hardware is a GeForce RTX 3090. IFED (f9) can achieve an inference speed of about \SI{24}{fps}. The quality performance in terms of PSNR/SSIM of each IFED variant calculated on different number of frames is listed in Table~\ref{tab:performance}. When the number of extracted frames is larger than 1, IFED (f9) is able to outperform other variants both in speed and accuracy. One possible reason for this is that training with more continuous ground-truth provides more time continuity to support learning.

\begin{table*}[!t] 	
	\begin{center}
		\resizebox{\linewidth}{!}{
			\begin{tabular}{@{\extracolsep{3pt}}lcccc@{}}
				\toprule
				& IFED (f1) & IFED (f3) & IFED (f5) & IFED (f9) \\
				\midrule
				Parameters & \SI{28.38}{M} & \SI{28.75}{M} & \SI{29.12}{M} & \SI{29.86}{M}\\
				Runtime & \SI{15.11}{ms} / \SI{8.70}{fps} & \SI{7.26}{ms} / \SI{13.77}{fps} & \SI{5.95}{ms} / \SI{16.81}{fps} & \SI{4.16}{ms} / \SI{24.04}{fps} \\
				\bottomrule
			\end{tabular}
		}
	\end{center}
	\caption{\textbf{The average inference cost for one frame (960$\times$540) of IFED.} f\# denotes the number of frames extracted by the model in a single inference.}
	\label{tab:inference_time}
\end{table*}

\begin{table*}[!t]	
	\begin{center}
		\begin{tabular}{@{\extracolsep{3pt}}lcccc@{}}
			\toprule
			& IFED (f1) & IFED (f3) & IFED (f5) & IFED (f9) \\
			\midrule
			1 frame & 32.07 / 0.934 & 30.99 / 0.915 & 31.21 / 0.919 & 31.11 / 0.920 \\
			3 frames & - & 28.48 / 0.87 & 28.66 / 0.876 & 28.70 / 0.880 \\
			5 frames & - & - & 29.79 / 0.897 & 29.93 / 0.901 \\
			9 frames & - & - & - & 30.34 / 0.910 \\
			\bottomrule
		\end{tabular}
	\end{center}
	\caption{\textbf{Performance of different variants of IFED in terms of PSNR/SSIM.}}
	\label{tab:performance}
\end{table*}

\subsection{Synchronization of Dual-RS Cameras}
\begin{figure}[!t]
	\centering
	\includegraphics[width=\linewidth]{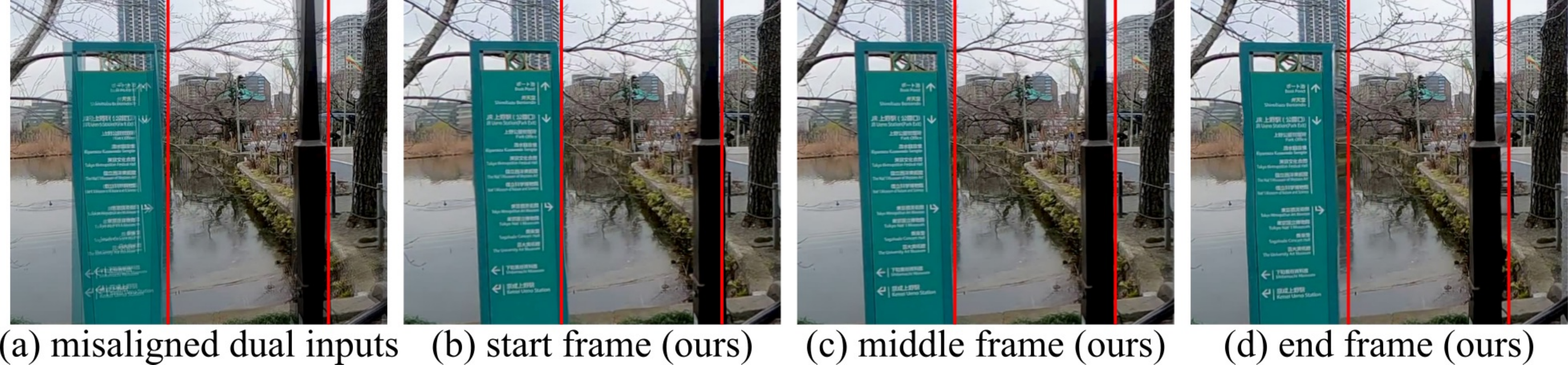}
	\caption{\textbf{Effectiveness of IFED with clock misalignment.}}
	\label{fig:clock}
\end{figure}

Today's clock synchronization circuits usually have errors below \SI{10}{\us}. In Fig.~\ref{fig:clock}, we show that our method is still effective when frames are misaligned by two rows to simulate out of synchronization.

\subsection{3D Reconstruction Evaluation}
\label{sec:reconstruction}
\begin{figure}[!t]
	\centering
	\includegraphics[width=\linewidth]{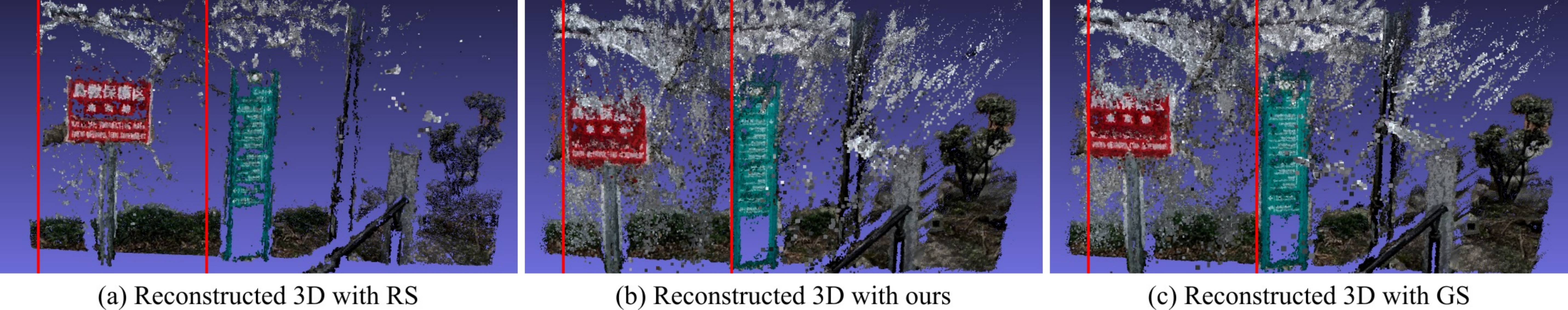}
	\caption{\textbf{Reconstruction performance.} 3D model from our corrected and interpolated RS images is closer to the model from GS.}
	\label{fig:sfm}
\end{figure}

In this section, we present the evaluation of method on 3D reconstruction task. We implemented SfM (\href{https://github.com/mapillary/OpenSfM}{OpenSfM}) to generate 3D models using images in the presence of camera rotations and translations as illustrated in Fig.~\ref{fig:sfm}. The structure built by using our RS corrected and interpolated images is closer to the one built by GS image sequence, which validates that our method can further serve to high-level tasks.

\subsection{Ablation Studies for Dual-RS Setup}
\begin{figure}[!t]
	\centering
	\begin{minipage}[h]{0.46\textwidth}
		\centering
		\includegraphics[width=\textwidth]{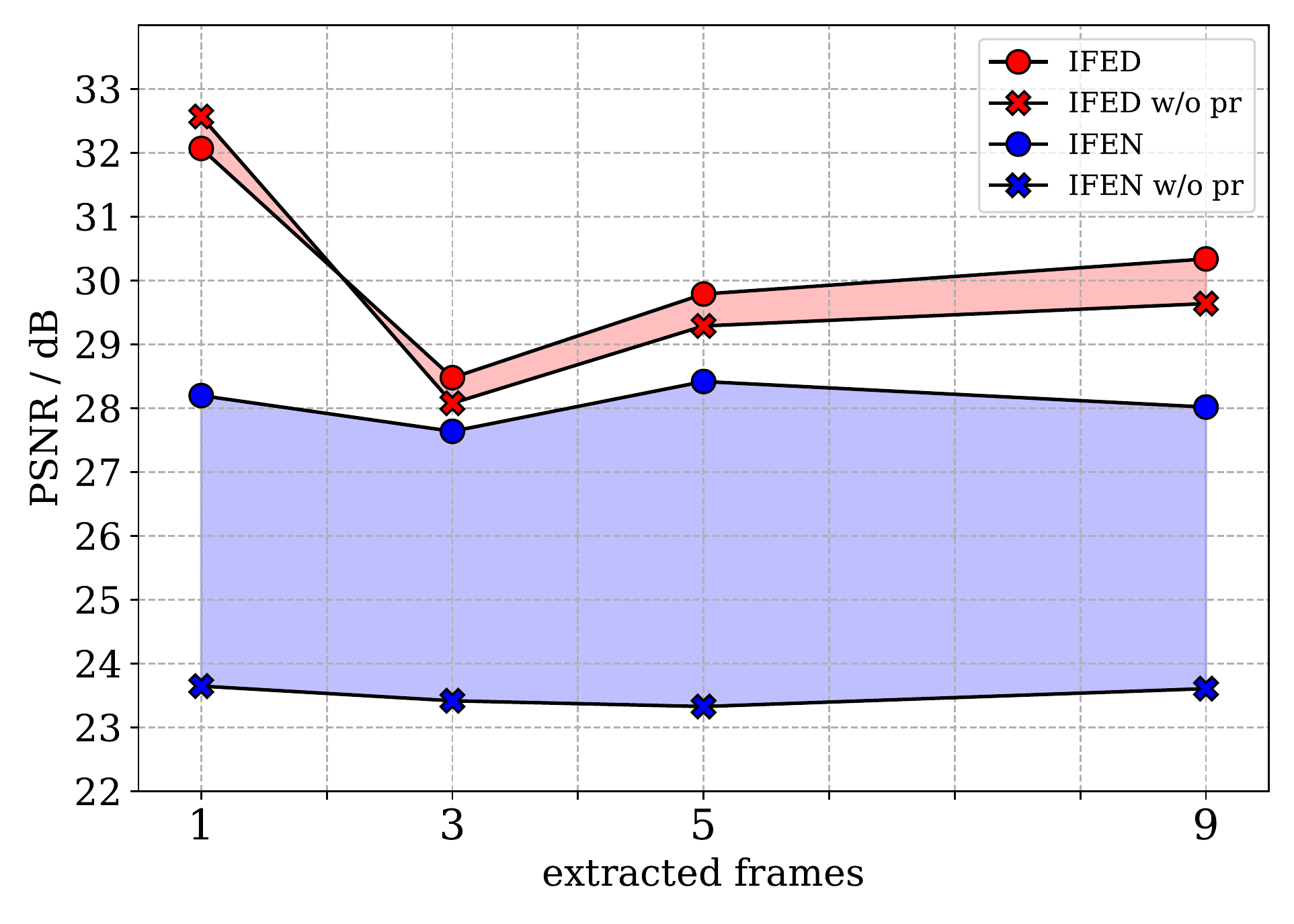}
		\caption{\textbf{Effectiveness of dual-RS setup and time cube prior.}}
		\label{fig:ablation}
	\end{minipage}
	\hfill
	\begin{minipage}[h]{0.50\textwidth}
		\centering
		\includegraphics[width=\textwidth]{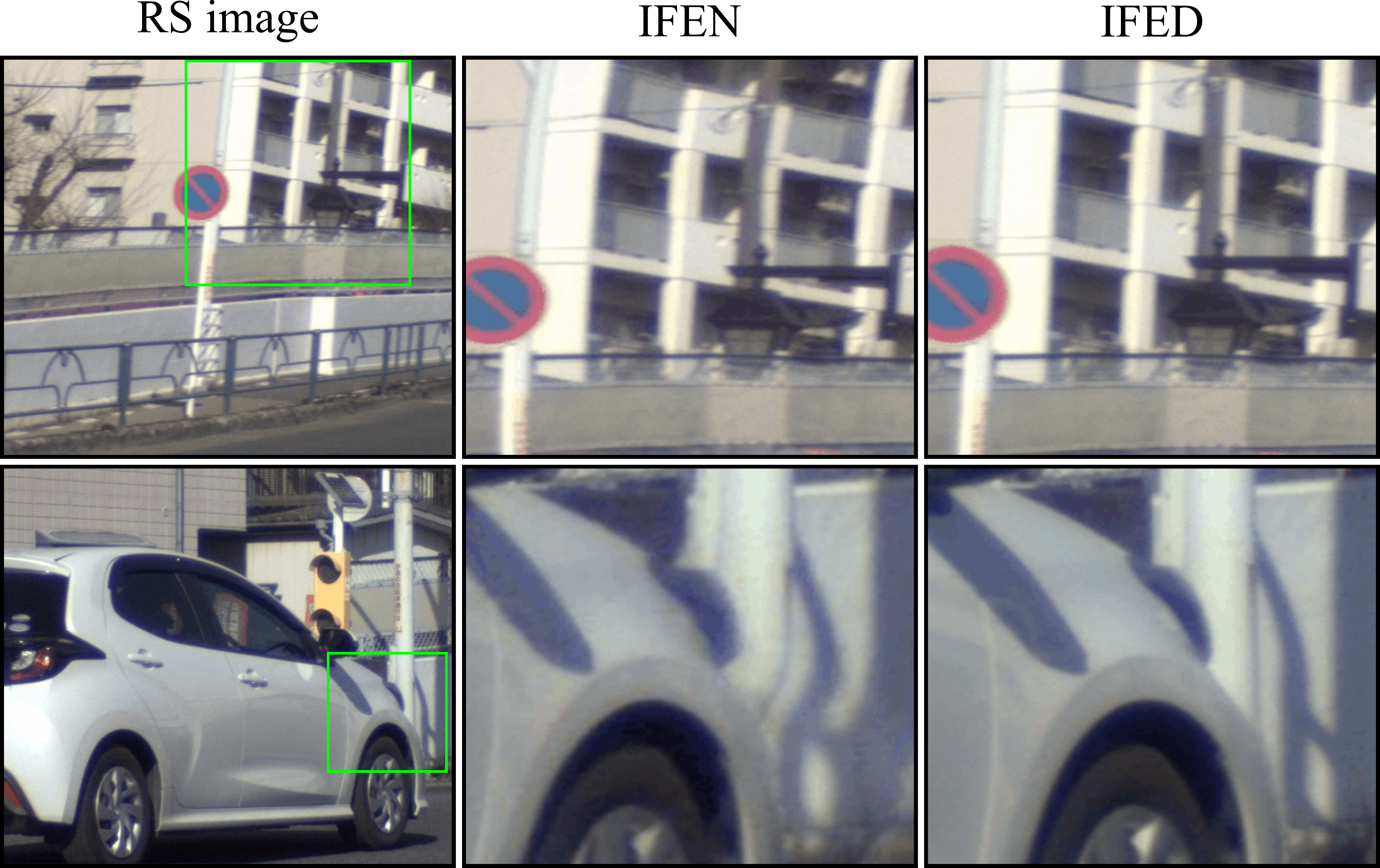}
		\caption{\textbf{The merit of dual-RS setup on real data.}}
		\label{fig:ifen}
	\end{minipage}
\end{figure}

We further show experiments of training our network using consecutive frames on RS-GOPRO, denoted as IFEN. The results also include whether to use time cube prior, as shown in Fig.~\ref{fig:ablation}. We can see dual-RS setup bring huge performance gain, and the time cube prior is more helpful in consecutive frame setup. When testing on real images with inconsistent readout settings with the training dataset, the advantage of dual-RS further extends because it avoids the undesired distortion caused by ambiguity, as shown in Fig.~\ref{fig:ifen}.

\section{Limitations}
We have succeeded in handling an electric fan with a spinning speed of up to \SI{500}{rpm}. However, it is likely to fail for larger objects at the same angular velocity, or smaller objects with faster angular velocity, as both have a faster linear speed around the outer edges.

\end{document}